\documentclass[11pt]{article} 

\usepackage{latexsym}

\title{Probabilistic Parsing Strategies}
\author{
\begin{tabular}[t]{c}
Mark-Jan Nederhof%
  \,\thanks{
Supported by the Royal Netherlands
Academy of Arts and Sciences.
Secondary affiliation is the
German Research Center for Artificial Intelligence (DFKI).
} \\
Faculty of Arts \\
University of Groningen \\
P.O.\ Box 716 \\
NL-9700 AS Groningen, The Netherlands \\
{\tt markjan@let.rug.nl}
\end{tabular}
\and
\begin{tabular}[t]{c}
Giorgio Satta \\
Department of Information Engineering
University of Padua \\
via Gradenigo, 6/A \\
I-35131 Padova, Italy \\
{\tt satta@dei.unipd.it}
\end{tabular}
}

\date{}

\input{epsf}

\newcommand{\comment}[1]{} 

\newcommand{\order}[1]{{\cal O}({#1})}

\newcommand{\mygram}{{\cal G}}
\newcommand{\myaut}{{\cal A}}
\newcommand{\mystrat}{{\cal S}}

\newcommand{\mypartial}{{\cal T}_{\myaut}}

\newcommand{\myterm}{\mit\Sigma}
\newcommand{\mynontset}{\mit\Gamma}
\newcommand{\mynont}{N}
\newcommand{\myrule}{R}

\newcommand{\bul}{\mathrel{\bullet}} 

\newcommand{\mysym}{Q}
\newcommand{\Xinit}{X_{\it init}}
\newcommand{\Xfinal}{X_{\it final}}
\newcommand{\mytrans}{\mit\Delta}

\newcommand{\myep}[2]{{#1} \mapsto {#2}}
\newcommand{\myscan}[4]{{#1} \stackrel{#2,#3}{\mapsto} {#4}}
\newcommand{\myscanrec}[3]{{#1} \stackrel{#2}{\mapsto} {#3}}

\newcommand{\pdamoverel}{\vdash}
\newcommand{\pdamove}[1]{\stackrel{#1}{\vdash}}
\newcommand{\pdamoves}{\vdash^\ast}
\newcommand{\pdamovesname}[1]{\stackrel{#1}{\vdash^\ast}}
\newcommand{\outp}{{\it out}}

\newcommand{\pdagoto}{\leadsto}

\newcommand{\de}{\rightarrow}

\newcommand{\LC}{\angle}
\newcommand{\LCep}{\angle_{\epsilon}}
\newcommand{\LCstar}{\angle^\ast}
\newcommand{\LCepstar}{\angle_{\epsilon}^\ast}

\newcommand{\fepLC}{f_{\epsilon\mbox{\scriptsize\it -LC}}}
\newcommand{\fepTD}{f_{\epsilon\mbox{\scriptsize\it -TD}}}
\newcommand{\stratepLC}{\mystrat_{\epsilon\mbox{\scriptsize\it -LC}}}

\newtheorem{theorem}{Theorem}
\newtheorem{lemma}[theorem]{Lemma}

\newcommand{\proof}{\noindent {\em Proof.\hspace{1em}}}
\newcommand{\closeproof}{\mbox{\hspace{1em}\rule{.45em}{.45em}}}

\newcommand{\tabrule}[4]{
\begin{eqnarray}
\label{#1}
        \frac{ \begin{array}{c} #2 \end{array} }
                        { \begin{array}{c} #3 \end{array} }
   \left\{ \begin{array}{l} #4 \end{array} \right.
\end{eqnarray} }

\newcommand{\tabruletwo}[3]{
\begin{eqnarray}
\label{#1}
        \frac{ \begin{array}{c}  #2 \end{array} }
                        { \begin{array}{c} #3 \end{array} }
\end{eqnarray} }

\newcommand{\forward}{{\it forward\/}}
\newcommand{\inner}{{\it inner\/}}
\newcommand{\tabel}{{\it tab\/}}

\begin{document}

\maketitle

\begin{abstract}
We present new results on the relation between 
purely symbolic context-free
parsing strategies and their probabilistic counter-parts.
Such parsing strategies are seen as constructions
of push-down devices from grammars.
We show that preservation of probability distribution is
possible under two conditions, viz.\
the correct-prefix property and the
property of strong predictiveness.
These results generalize existing results in the literature 
that were obtained by considering parsing strategies in 
isolation.  From our general results we also derive negative
results on so-called generalized LR parsing. 
\end{abstract}

\section{Introduction}
\label{s:intro}

Context-free grammars and push-down automata are two 
equivalent formalisms to describe context-free languages.
While a context-free grammar can be thought of as a purely
declarative specification, a push-down automaton is considered to be
an operational specification that determines which steps are 
performed for a given string
in the process of deciding its membership of the language.
By a {\em parsing strategy\/} we mean a mapping from 
context-free grammars to equivalent push-down automata, 
such that some specific conditions are observed. 

This paper deals with the probabilistic extensions of 
context-free grammars and push-down automata, 
i.e., probabilistic context-free grammars \cite{SA72,BO73}
and probabilistic push-down automata \cite{SA72,SA76,TE95,AB99}.
These formalisms
are obtained by adding probabilities to the rules and transitions
of context-free grammars and push-down automata, respectively.
More specifically, we will investigate the problem of `extending'
parsing strategies to {\em probabilistic\/} parsing strategies.
These are mappings from probabilistic context-free grammars to 
probabilistic push-down automata
that preserve the induced probability distributions
on the generated/accepted languages.
Two of the main results presented in this paper can be stated as follows:
\begin{itemize}
\item 
No parsing strategy that lacks the
correct-prefix property (CPP) can be extended to become
a probabilistic parsing strategy.
\item
All parsing strategies that possess
the correct-prefix property and the 
strong predictiveness property (SPP) can be extended
to become probabilistic parsing strategies.
\end{itemize}
The above results generalize previous findings
reported in~\cite{TE95,TE97,AB99}, where only a few specific 
parsing strategies were considered in isolation. 
Our findings also have important
implications for well-known parsing strategies such as
generalized LR parsing, henceforth simply called `LR parsing'.%
\footnote{Generalized (or nondeterministic)
LR parsing allows for more than one action
for a given LR state and input symbol.} 
LR parsing has the CPP, but lacks the SPP, and as we
will show, LR parsing cannot be extended to become
a probabilistic parsing strategy.

In the last decade, widespread interest
in probabilistic parsing techniques has arisen
in the area of natural language processing \cite{CH93a,MA99,JU00}.
This is motivated by the fact that natural language sentences are
generally ambiguous,
and natural language software needs to be able to
distinguish the more probable derivations
of a sentence from the less probable ones.
This can be achieved by letting the parsing process
assign a probability to each parse, 
on the basis of a probabilistic grammar.
In a typical application, the software may 
select those derivations for further processing
that have been given the highest
probabilities, and discard the others. 
The success of this approach relies on the accuracy of
the probabilistic model expressed by the probabilistic
grammar, i.e., whether the probabilities assigned
to derivations accurately reflect
the `true' probabilities in the domain at hand.

Probabilities are often estimated on the basis of a corpus,
i.e., a collection of sentences. The sentences in a corpus 
may be annotated with various kinds of information. 
One kind of annotation that is relevant for our discussion is
the preferred derivation for each sentence. 
Given a corpus with derivations, one may
estimate probabilities of rules by their relative
frequencies in the corpus. If a corpus is unannotated, 
more general techniques of maximum-likelihood estimation
can be used to estimate the probabilities of rules. 
(See \cite{SA97,CH98,CH99} for some formal properties of types of
maximum-likelihood estimation.)

The motivation for studying probabilistic models other than
those obtained by attaching probabilities to 
given context-free grammars is the
observation that more accurate models can be obtained by 
conditioning probabilities on `context information' beyond single
nonterminals \cite{CH90,CH94}. Furthermore, it
has been observed that conditioning on certain types of
context information can be achieved by first translating
context-free grammars to push-down automata,
according to some parsing strategy,
and then attaching probabilities to the transitions thereof 
\cite{SO99,RO99}. 
More concretely, for some parsing strategies,
the set of models that can be obtained by attaching
probabilities to a push-down automaton constructed from
a context-free grammar may include models that cannot be obtained by
attaching probabilities to that grammar.

An implicit assumption of this methodology is that,
conversely, any probabilistic model that can be obtained from a grammar
can also be obtained from the associated push-down automaton,
or in other words, the push-down automaton is at least as
powerful as the grammar in terms of the set the potential models.
If a parsing strategy does not satisfy this property, and
if some potential models are lost in the mapping from
the grammar to the push-down automaton, then this means that
in some cases
the strategy may lead to less rather than more accurate models.
That LR parsing cannot be extended to become
a probabilistic parsing strategy, as we mentioned above,
means that the above property is not satisfied by this parsing strategy.
This is contrary to what is suggested by some
publications on probabilistic LR parsing, such as
\cite{BR93} and \cite{IN00}, which fail to observe that
LR parsers may sometimes lead to less accurate models
than the grammars from which they were constructed.

Some studies, such as \cite{CO97,CH98a,CH01}, propose
lexicalized probabilistic context-free grammars, i.e., 
probabilistic models based on 
context-free grammars in which probabilities
heavily rely on the terminal elements from input strings.
Even if the current paper does not specifically deal with
lexicalization, much of what we discuss pertains
to lexicalized probabilistic context-free grammars as well.

The paper is organized as follows. After giving standard
definitions in Section~\ref{s:prel}, we give our formal
definition of `parsing strategy' in Section~\ref{s:strategy}.
We also define what it means to extend a parsing strategy to
become a probabilistic parsing strategy.
The CPP and the SPP are defined in Sections~\ref{s:cpp}
and~\ref{s:pred}, where we also discuss how these properties relate to
the question of which strategies can be extended to become
probabilistic.
Sections~\ref{s:strong} 
and~\ref{s:nonstrong} provide examples of parsing strategies
with and without the SPP. The examples without the SPP, 
most notably LR parsing, are
shown not to be extendible to become probabilistic.
A wider notion of extending a strategy to become probabilistic
is provided by Section~\ref{s:wide}. We show that
even under this wider notion,
LR parsing cannot be extended to become probabilistic.
Section~\ref{s:prefix} presents an application
that concerns prefix probabilities.
We end this paper with conclusions.

Some results reported here have appeared before in an abbreviated form
in \cite{NE02d}.

\section{Preliminaries}
\label{s:prel}

A context-free grammar (CFG) $\mygram$ is a 4-tuple 
$(\myterm,$ $\mynont,$ $S,$ $\myrule)$,
where $\myterm$ is a finite set of {\em terminals}, 
called the {\em alphabet},
$\mynont$ is a finite set of {\em nonterminals},
including the {\em start symbol\/} $S$, and $\myrule$ is a finite set of
{\em rules},
each of the form $A\de\alpha$, where $A\in \mynont$ and
$\alpha\in (\myterm \cup \mynont)^\ast$.
Without loss of generality, we assume that there is only one
rule $S \de \sigma$ with the start symbol in the left-hand side,
and furthermore that $\sigma \neq \epsilon$, where $\epsilon$
denotes the empty string.

For a fixed CFG $\mygram$, we
define the relation $\Rightarrow$ on triples consisting of two strings
$\alpha,\beta\in (\myterm \cup \mynont)^\ast$ and a rule 
$\pi\in\myrule$ by:
$\alpha \stackrel{\pi}{\Rightarrow} \beta$ if and only if
$\alpha$ is of the form $wA\delta$ and $\beta$ is of the
form $w\gamma\delta$, for some $w\in\myterm^\ast$ and
$\delta\in (\myterm \cup \mynont)^\ast$, and $\pi=(A\de\gamma)$.
A {\em left-most derivation\/} 
is a string $d = \pi_1 \cdots \pi_m$, $m \geq 0$,
such that $S \stackrel{\pi_1}{\Rightarrow} \cdots
\stackrel{\pi_m}{\Rightarrow}\alpha$, 
for some $\alpha \in (\myterm \cup \mynont)^\ast$.
We will identify a left-most derivation
with the sequence of strings over 
$\myterm \cup \mynont$ that arise in that
derivation. 
In the remainder of this paper, we will let
the term `derivation' refer to
`left-most derivation', unless specified otherwise.

A derivation $d = \pi_1 \cdots \pi_m$, $m \geq 0$,
such that $S \stackrel{\pi_1}{\Rightarrow} \cdots
\stackrel{\pi_m}{\Rightarrow}w$ where $w\in\myterm^\ast$ 
will be called a {\em complete\/} derivation;
we also say that $d$ is a derivation of $w$.
By {\em subderivation\/} we mean a substring of a
complete derivation of the form
$d = \pi_1 \cdots \pi_m$, $m \geq 0$,
such that $A \stackrel{\pi_1}{\Rightarrow} \cdots
\stackrel{\pi_m}{\Rightarrow}w$ for some $A$ and $w$.

We write $\alpha \Rightarrow^\ast \beta$ or
$\alpha \Rightarrow^+ \beta$ to denote the
existence of a string $\pi_1 \cdots \pi_m$ such that
$\alpha \stackrel{\pi_1}{\Rightarrow} \cdots
\stackrel{\pi_m}{\Rightarrow}\beta$,
with $m \geq 0$ or $m > 0$, respectively.
We say a CFG is {\em acyclic\/} if
$A \Rightarrow^+ A$ does not hold for any $A\in\mynont$.

For a CFG $\mygram$ we define the language $L(\mygram)$
it generates as the set of strings $w$
such that there is at least one derivation of $w$.
We say a CFG is {\em reduced\/} if for each rule $\pi\in\myrule$
there is a complete derivation in which it occurs.

A {\em probabilistic\/} context-free grammar (PCFG) is a pair
$(\mygram, p)$ consisting of a CFG 
$\mygram=(\myterm,$ $\mynont,$ $S,$ $\myrule)$ and 
a probability function $p$ from $\myrule$ to
real numbers in the interval $[0,1]$. We say a PCFG is {\em proper\/}
if $\Sigma_{\pi=(A\de\gamma)\in\myrule}\ p(\pi) = 1$ for
each $A\in\mynont$.

For a PCFG $(\mygram,p)$,
we define
the probability $p(d)$ of a string 
$d = \pi_1 \cdots \pi_m \in \myrule^\ast$
as $\prod_{i=1}^m\  p(\pi_i)$; 
we will in particular consider the probabilities of
derivations $d$.
The probability $p(w)$ of a string $w\in \myterm^\ast$ as defined by $(\mygram,p)$
is the sum of the probabilities of
all derivations of that string.
We say a PCFG $(\mygram,p)$ is {\em consistent\/} if
$\Sigma_{w \in \myterm^\ast}\ p(w) = 1$.

In this paper we will mainly consider push-down transducers
rather than push-down automata. Push-down transducers not
only compute derivations of the grammar while processing
an input string, but they also explicitly produce 
output strings from which these derivations can be obtained. 
We use transducers for two reasons.
First, constraints on the output strings allow
us to restrict our attention to `reasonable' parsing strategies.
Those strategies that cannot be formalized within these constraints
are unlikely to be of practical interest.
Secondly, mappings from input strings to derivations, as those realized 
by push-down devices, turn out to be a very powerful abstraction 
and allow direct proofs of several general results.  

Differently from many textbooks, our push-down devices do not
possess states next to stack symbols. This is without loss
of generality, since states can be encoded into the stack symbols,
given the types of transition that we allow.
Thus,
a push-down transducer (PDT) $\myaut$ is a 6-tuple
$(\myterm_1,$ $\myterm_2,$ $\mysym,$ $\Xinit,$ $\Xfinal,$ $\mytrans)$,
where $\myterm_1$ is the input alphabet,
$\myterm_2$ is the output alphabet,
$\mysym$ is a finite set of {\em stack symbols}
including the {\em initial stack symbol\/} $\Xinit$ and the
{\em final stack symbol\/} $\Xfinal$, and $\mytrans$ is the set of
{\em transitions}.
Each transition can have
one of the following three forms:
$\myep{X}{X Y}$ (a push transition),
$\myep{\it Y X}{Z}$ (a pop transition),  or
$\myscan{X}{x}{y}{Y}$ (a swap transition);
here $X$, $Y$, $Z\in \mysym$,
$x\in \myterm_1 \cup \{\epsilon\}$ 
and $y\in \myterm_2^\ast$.
Note that
in our notation, stacks grow from left to right, i.e., the top-most
stack symbol will be found at the right end.

Without loss of generality, we assume that any PDT is such that
for a given stack symbol $X\neq \Xfinal$, there are either one or more
push transitions
$\myep{X}{X Y}$, or one or more pop transitions 
$\myep{\it Y X}{Z}$, or one or more swap transitions
$\myscan{X}{x}{y}{Y}$, but no combinations of different types of
transition. If a PDT does not satisfy this normal form, it can
easily be brought in this form by introducing for each stack symbol
$X$ three new stack symbols $X_{\it push}$, $X_{\it pop}$
and $X_{\it swap}$ and new swap transitions 
$\myscan{X}{\epsilon}{\epsilon}{X_{\it push}}$,
$\myscan{X}{\epsilon}{\epsilon}{X_{\it pop}}$ and
$\myscan{X}{\epsilon}{\epsilon}{X_{\it swap}}$.
In each existing transition that operates on top-of-stack $X$,
we then replace $X$ by one from $X_{\it push}$, $X_{\it pop}$
or $X_{\it swap}$, depending on the type of that transition.
We also assume that $\Xfinal$ does not occur in the left-hand side
of a transition, again without loss of generality.

A {\em configuration\/} of a PDT is a triple
$(\alpha, w, v)$, where $\alpha \in \mysym^\ast$
is a stack, $w\in\myterm_1^\ast$ is the remaining input, and
$v\in\myterm_2^\ast$ is the output generated so far.
For a fixed PDT $\myaut$, we define
the relation $\pdamoverel$ on triples consisting of two
configurations and a transition $\tau$ by:
$(\gamma\alpha, xw, v) \pdamove{\tau} (\gamma\beta, w, vy)$ if and only if
$\tau$ is of the form
$\myep{\alpha}{\beta}$, where $x=y=\epsilon$, or of the form
$\myscan{\alpha}{x}{y}{\beta}$.
A {\em computation\/} on an input string $w$ is a string
$c=\tau_1 \cdots \tau_m$, $m \geq 0$, such that
$(\Xinit, w, \epsilon) \pdamove{\tau_1} \cdots \pdamove{\tau_m}
(\alpha, w', v)$.
A {\em complete\/} computation on a string $w$ is a computation
with $w'=\epsilon$ and $\alpha=\Xfinal$. The string $v$ is called
the {\em output\/} of the computation $c$, 
and is denoted by $\outp(c)$.

We will identify a
computation with the sequence of configurations
that arise in that computation,
where the first configuration is determined by the context.
We also write
$(\alpha,w,v) \pdamoves (\beta,w',v')$ or 
$(\alpha,w,v) \pdamovesname{c} (\beta,w',v')$, 
for $\alpha,\beta \in \mysym^\ast$, 
$w,w'\in \myterm_1^\ast$ and $v,v'\in\myterm_2^\ast$,
to indicate that $(\beta,w',v')$ can be obtained
from $(\alpha,w,v)$ by applying a sequence $c$ of zero or more
transitions; we refer to such a sequence $c$ 
as a {\em subcomputation}. 
The function $\outp$ is
extended to subcomputations in a natural way.

For a PDT $\myaut$, we define the language $L(\myaut)$ it
accepts as the set of strings $w$ such that there is
at least one complete computation on $w$.
We say a PDT is {\em reduced\/} if 
each transition $\tau\in\mytrans$ occurs in some complete computation.

A {\em probabilistic\/} push-down transducer (PPDT) is a pair
$(\myaut, p)$ consisting of a PDT $\myaut$ and
a probability function $p$ from the set $\mytrans$ of
transitions of $\myaut$ to
real numbers in the interval $[0,1]$.
We say a PPDT $(\myaut, p)$ is {\em proper\/} if
\begin{itemize}
\item 
$\Sigma_{\tau=(\myep{X}{X Y})\in\mytrans}\ p(\tau) = 1$
for each $X\in\mysym$ such that there is at least one
transition $\myep{X}{X Y}$, $Y \in \mysym$;
\item
$\Sigma_{\tau=(\myscan{X}{x}{y}{Y})\in\mytrans}\ p(\tau) = 1$
for each $X\in\mysym$ such that there is at least one
transition $\myscan{X}{x}{y}{Y}$, 
$x\in\myterm_1\cup\{\epsilon\},y\in\myterm_2^\ast,Y\in\mysym$; 
and
\item
$\Sigma_{\tau=(\myep{Y X}{Z})\in\mytrans}\ p(\tau) = 1$,
for each $X,Y\in\mysym$ such that there is at least one
transition $\myep{Y X}{Z}$, $Z\in\mysym$.
\end{itemize}

For a PPDT $(\myaut,p)$,
we define the probability $p(c)$
of a (sub)computation $c=\tau_1 \cdots \tau_m$ 
as $\prod_{i=1}^m\  p(\tau_i)$.
The probability $p(w)$ of a string $w$ as defined by $(\myaut,p)$
is the sum of the probabilities of
all complete computations on that string.
We say a PPDT $(\myaut,p)$ is {\em consistent\/} if
$\Sigma_{w \in \myterm^\ast}\ p(w) = 1$.

We say a PCFG $(\mygram,p)$ is reduced if $\mygram$ is reduced,
and we say a PPDT $(\myaut,p)$ is reduced if $\myaut$ is reduced.

\section{Parsing strategies}
\label{s:strategy}

The term `parsing strategy' is often used informally to
refer to a class of parsing algorithms that behave similarly
in some way. In this paper, we assign a formal
meaning to this term, relying on the 
observation by \cite{LA74,BI89} that many
parsing algorithms for CFGs can be described in two steps.
The first is a construction of push-down devices 
from CFGs, and the second is
a method for handling nondeterminism 
(e.g.\ backtracking or dynamic programming).
Parsing algorithms that handle nondeterminism in
different ways but apply the same construction of
push-down devices from CFGs are seen as realizations of
the same parsing strategy.

Thus, we define a {\em parsing strategy\/} to be a function 
$\mystrat$ that maps 
a reduced CFG $\mygram
=(\myterm_1,$ $\mynont,$ $S,$ $\myrule)$ to
a pair $\mystrat(\mygram)=(\myaut,f)$ consisting of a 
reduced PDT $\myaut=(\myterm_1,$ $\myterm_2,$ $\mysym,$ 
$\Xinit,$ $\Xfinal,$ $\mytrans)$, and a function $f$ that maps a subset of 
$\myterm_2^\ast$ to a subset of $\myrule^\ast$,
with the following properties:
\begin{itemize}
\item $\myrule \subseteq \myterm_2$.
\item For each string $w\in\myterm_1^\ast$ and each
complete computation $c$ on $w$,
$f(\outp(c))=d$ is a derivation of $w$.
Furthermore, each symbol from $\myrule$
occurs as often in $\outp(c)$ as it occurs in $d$.
\item Conversely, for each string $w\in\myterm_1^\ast$ and 
each derivation $d$ of $w$,
there is precisely one complete computation $c$ on $w$ such that
$f(\outp(c)) = d$.
\end{itemize}
If $c$ is a complete computation, we will write
$f(c)$ to denote $f(\outp(v))$. The conditions
above then imply that $f$ is
a bijection from complete computations to complete derivations.

Note that output strings
of (complete) computations may contain symbols that are not in $\myrule$,
and the symbols that are in $\myrule$ may occur in a different
order in $v$ than in $f(v)=d$. The purpose of the symbols
in $\myterm_2 - \myrule$ is to help this process of reordering
of symbols in $\myrule$.
For a string $v \in \myterm_2^\ast$ we let $\overline{v}$ refer
to the maximal subsequence of symbols from $v$ that belong to $\myrule$,
or in other words, string $\overline{v}$ is obtained by erasing
from $v$ all occurrences of symbols from $\myterm_2 - \myrule$.

A {\em probabilistic parsing strategy\/} is defined to be a function
$\mystrat$ that maps a reduced, proper and consistent
PCFG $(\mygram, p_{\mygram})$ 
to a triple $\mystrat(\mygram,  p_{\mygram})=(\myaut, p_{\myaut}, f)$,
where $(\myaut, p_{\myaut})$ is a reduced, proper and consistent PPDT,
with the same properties as a
(non-probabilistic) parsing strategy, and in addition:
\begin{itemize}
\item
For each complete derivation $d$ and
each complete computation $c$ such that $f(c)=d$,
$p_{\mygram}(d)$ equals $p_{\myaut}(c)$.
\end{itemize}
In other words, a complete computation has the same probability
as the complete derivation that it is mapped to by
function $f$.
An implication of this property is that for each string $w\in\myterm_1^\ast$,
the probabilities assigned to that string
by $(\mygram, p_{\mygram})$ and $(\myaut,p_{\myaut})$ are equal.

We say that probabilistic parsing strategy $\mystrat'$ 
is an {\em extension\/} of parsing strategy $\mystrat$ if
for each reduced CFG $\mygram$ and probability function $p_{\mygram}$
we have
$\mystrat(\mygram)=(\myaut, f)$ if and only if
$\mystrat'(\mygram, p_{\mygram})=(\myaut, p_{\myaut}, f)$
for some $p_{\myaut}$.

In the following sections we will investigate which
parsing strategies can be extended to become
probabilistic parsing strategies.

\section{Correct-prefix property}
\label{s:cpp}

For a given PDT,
we say a computation $c$ is {\em dead\/} if 
$(\Xinit, w_1, \epsilon)$ $\pdamovesname{c}$
$(\alpha, \epsilon, v_1)$, for some $\alpha\in\mysym^\ast$, 
$w_1\in \myterm_1^\ast$ and $v_1\in\myterm_2^\ast$,
and there are no
$w_2\in \myterm_1^\ast$ and $v_2\in\myterm_2^\ast$ such that
$(\alpha, w_2, \epsilon) \pdamoves (\Xfinal, \epsilon, v_2)$.
Informally, a dead computation is a computation that
cannot be continued to become a complete computation.

We say that a PDT has the {\em correct-prefix property\/} (CPP) if
it does not allow any dead computations.
We say that a parsing strategy has the CPP if it maps each
reduced CFG to a PDT that has the CPP.

In this section we show that the correct-prefix property is 
a necessary condition
for extending a parsing strategy to a probabilistic parsing strategy.
For this we need two lemmas.

\begin{lemma}
\label{l:pcfg}
For each reduced CFG $\mygram$, there is a probability function
$p_{\mygram}$ such that 
PCFG $(\mygram,p_{\mygram})$ is proper and consistent,
and $p_{\mygram}(d) > 0$ for all complete derivations~$d$.
\end{lemma}

\proof
Since $\mygram$ is reduced, there is a finite set $L$ consisting of
complete derivations $d$, such that for each rule $\pi$ in $\mygram$
there is at least
one $d\in L$ in which $\pi$ occurs.
Let $n_{\pi,d}$ be the number of occurrences of rule $\pi$ in
derivation $d\in L$, and let $n_{\pi}$ be
$\Sigma_{d\in L}\ n_{\pi,d}$, 
the total number of occurrences of $\pi$ in $L$.
Let $n_A$ be the sum of $n_{\pi}$ for all rules 
$\pi$ with $A$ in the left-hand side. A probability function
$p_{\mygram}$ can be defined through
`maximum-likelihood estimation' such that
$p_{\mygram}(\pi) = \frac{n_{\pi}}{n_A}$ for each rule
$\pi = A \de \alpha$. 

For all nonterminals $A$, 
$\Sigma_{\pi = A \de \alpha}\ p_{\mygram}(\pi)$ $=$
$\Sigma_{\pi = A \de \alpha}\ \frac{n_{\pi}}{n_A} $=$ \frac{n_A}{n_A}$ $=$ 1,
which means that the PCFG $(\mygram,p_{\mygram})$ is proper.
Furthermore, \cite{CH98} has shown that a PCFG $(\mygram,p_{\mygram})$ 
is consistent if
$p_{\mygram}$ was obtained by maximum-likelihood estimation using 
a set of derivations.
Finally, since $n_{\pi} > 0$ for each $\pi$, also
$p_{\mygram}(\pi) > 0$ for each $\pi$, and
$p_{\mygram}(d) > 0$ for all complete derivations $d$.~\closeproof

We say a computation is a {\em shortest\/} dead computation if
it is dead and none of its proper prefixes is dead.
Note that each dead computation has a unique prefix that is a
shortest dead computation.
For a PDT $\myaut$, let $\mypartial$ be the union of the set of
all complete computations and the set of all
shortest dead computations.

\begin{lemma}
\label{l:partial}
For each proper PPDT $(\myaut, p_{\myaut})$, 
$\Sigma_{c \in \mypartial}\  p_{\myaut}(c) \leq 1$.
\end{lemma}

\proof
The proof is a trivial variant of the proof
that for a proper PCFG $(\mygram, p_{\mygram})$,
the sum of $p_{\mygram}(d)$ for all derivations $d$ cannot
exceed 1, which is shown by \cite{BO73}.~\closeproof

{}From this, the main result of this section follows.

\begin{theorem}
\label{t:cpp}
A parsing strategy that lacks the CPP cannot be extended to 
become a probabilistic parsing strategy.
\end{theorem}

\proof
Take a parsing strategy $\mystrat$ that does not have the CPP. 
Then there is a reduced
CFG $\mygram= (\myterm_1,$ $\mynont,$ $S,$ $\myrule)$, 
with $\mystrat(\mygram) = (\myaut,f)$ for some
$\myaut$ and $f$, and a shortest dead computation $c$ allowed 
by $\myaut$.

It follows from Lemma~\ref{l:pcfg} that there is a probability function
$p_{\mygram}$ such that
$(\mygram, p_{\mygram})$ is a proper and consistent PCFG and
$p_{\mygram}(d) > 0$ for all complete derivations $d$.
Assume we also have a probability function $p_{\myaut}$ such that
$(\myaut, p_{\myaut})$ is a proper and consistent PPDT
that assigns the same probabilities to strings over $\Sigma_1$ as
$(\mygram, p_{\mygram})$. Since $\myaut$ is reduced, each
transition $\tau$ must occur in some complete computation $c'$. Furthermore,
for each complete computation $c'$ there is a complete derivation $d$
such that $f(c') = d$, and $p_{\myaut}(c') = p_{\mygram}(d) > 0$. 
Therefore, $p_{\myaut}(\tau) > 0$ for each
transition $\tau$, and $p_{\myaut}(c) > 0$,
where $c$ is the above-mentioned dead computation.

Due to Lemma~\ref{l:partial},
$1 \geq \Sigma_{c' \in \mypartial}\  p_{\myaut}(c') \geq
\Sigma_{w \in \myterm_1^\ast}\ p_{\myaut}(w) + p_{\myaut}(c) >
\Sigma_{w \in \myterm_1^\ast}\ p_{\myaut}(w) = 
\Sigma_{w \in \myterm_1^\ast}\ p_{\mygram}(w)$.
This is in contradiction with the consistency of $(\mygram, p_{\mygram})$.
Hence, a probability function $p_{\mygram}$ with the properties we
required above cannot exist, and therefore $\mystrat$ cannot be extended 
to become
a probabilistic parsing strategy.~\closeproof

\section{Strong predictiveness}
\label{s:pred}

For a fixed PDT, we define the binary relation
$\pdagoto$ on stack symbols by:
$Y\pdagoto Y'$ if and only if 
$(Y, w, \epsilon) \pdamoves (Y',\epsilon, v)$ for some
$w \in \myterm_1^\ast$ and $v\in \myterm_2^\ast$. 
In other words,
some subcomputation may start with stack $Y$ and 
end with stack $Y'$. Note that all 
stacks that occur in such a subcomputation 
must have height of~1 or more.

We say that a PDT has the {\em strong predictiveness property\/} 
(SPP) if the existence of
three transitions $\myep{X}{X Y}$, $\myep{X Y_1}{Z_1}$ and
$\myep{X Y_2}{Z_2}$ such that $Y\pdagoto Y_1$ and
$Y\pdagoto Y_2$ implies $Z_1 = Z_2$. 
Informally, this means that
when a subcomputation starts with 
some stack $\alpha$ and some push transition $\tau$, 
then solely on the basis of $\tau$
we can uniquely determine
what stack symbol $Z_1 = Z_2$ will be on top of the stack in
the first configuration 
with stack height equal to $|\alpha|$. 
Another way of looking at
it is that no information may flow from higher stack elements
to lower stack elements that
was not already predicted before these higher stack elements
came into being, hence the term `strong predictiveness'.%
\footnote{There is a property of push-down devices called
{\em faiblement pr{\'e}dictif\/} (weakly predictive) \cite{VI93a}.
Contrary to what this name may suggest however, this property
is incomparable with the complement of our notion of SPP.}

We say that a parsing strategy has the SPP if it maps each
reduced CFG to a PDT with the SPP.

In the previous section it was shown that we may restrict ourselves
to parsing strategies that have the CPP. Here we show that
if, in addition, a parsing strategy has the SPP, then it can
always be extended to become a probabilistic parsing strategy.

\begin{theorem}
\label{t:sp}
Any parsing strategy that has the CPP and the SPP
can be extended to become a probabilistic parsing strategy.
\end{theorem}

\proof
Take a parsing strategy $\mystrat$ that 
has the CPP and the SPP,
and take a reduced PCFG $(\mygram,p_{\mygram})$,
where $\mygram = (\myterm_1,$ $\mynont,$  $S,$ $\myrule)$,   
and let $\mystrat(\mygram) = (\myaut,f)$, for some PDT $\myaut$ and
function $f$.
We will show that there is a probability function 
$p_{\myaut}$ such that $(\myaut, p_{\myaut})$ is a PPDT and
$p_{\myaut}(c) = p_{\mygram}(f(c))$ for all complete computations $c$.

For each stack symbol $X$, consider the set of transitions that
are applicable with top-of-stack $X$. Remember that our normal form
ensures that all such transitions are of the same type.
Suppose this set consists of $m$ swap transitions
${\tau_i} = \myscan{X}{x_i}{y_i}{Y_i}$, $1 \leq i \leq m$.
For each $i$,
consider all subcomputations of the form
$({\it X}, x_iw, \epsilon)$ $\pdamove{\tau_i}$ $({\it Y_i}, w, y_i)$
$\pdamoves$
$({\it Y'}, \epsilon, v)$ such that there is at least one
pop transition of the form $\myep{{\it Z Y'}}{Z'}$ or 
such that $Y' = \Xfinal$,
and define $L_{\tau_i}$ as the set of strings $v$
output by these subcomputations.
We also define $L_X = \cup_{j=1}^m\  L_{\tau_j}$, the set
of all strings output by subcomputations starting with top-of-stack 
$X$, and ending just before
a pop transition that leads to a stack with height smaller than that 
of the stack at the beginning, or ending with the final stack symbol $\Xfinal$.

Now define for each $i$ ($1 \leq i \leq m$):
\begin{eqnarray}
\label{e:normalized}
p_{\myaut}(\tau_i) &=&
\frac{  \Sigma_{v\in L_{\tau_i}}\ p_{\mygram}(\overline{v}) }{
        \Sigma_{v\in L_{X}}\ p_{\mygram}(\overline{v}) }
\end{eqnarray}
In other words, the probability of a transition is the normalized
probability of the set of subcomputations starting with that transition,
relating subcomputations with fragments of derivations of the PCFG.
 
These definitions are well-defined. Since $\myaut$ is reduced and has
the CPP, the sets $L_{\tau_i}$ are non-empty and thereby
the denominator in the definition of $p_{\myaut}(\tau_i)$
is non-zero. Furthermore,
$\Sigma_{i=1}^m\ p_{\myaut}(\tau_i)$ is clearly $1$.
 
Now suppose the set of transitions for $X$ consists of $m$
push transitions
${\tau_i} = \myep{X}{X Y_i}$, $1 \leq i \leq m$.
For each $i$,
consider all subcomputations of the form
$({\it X}, w, \epsilon)$ $\pdamove{\tau_i}$ $({\it XY_i}, w, \epsilon)$
$\pdamoves$
$({\it X'}, \epsilon, v)$ such that there is at least one
pop transition of the form $\myep{{\it Z X'}}{Z'}$ or $X' = \Xfinal$,
and define $L_{\tau_i}$, $L_X$ and $p_{\myaut}(\tau_i)$ as we have done
above for the swap transitions.
 
Suppose the set of transitions for $X$ consists of $m$
pop transitions 
${\tau_i} = \myep{\it Y_iX}{Z_i}$, $1 \leq i \leq m$. 
Define
$L_X = \{\epsilon\}$, and $p_{\myaut}(\tau_i)=1$ for each $i$.
To see that this is compatible with the condition of properness
of PPDTs,
note the following.
Since we may assume $\myaut$ is reduced,
if $Y_i = Y_j$ for some $i$ and $j$ with $1 \leq i,j \leq m$, 
then there is at least one
transition $\myep{Y_i}{\it Y_i X'}$ for some $X'$ such that
$X'\pdagoto X$. Due to the SPP, $Z_i = Z_j$ and therefore $i=j$.

Finally, we define $L_{\Xfinal} = \{\epsilon\}$.

Take a subcomputation $({\it X}, w, \epsilon)$
$\pdamovesname{c}$
$({\it Y}, \epsilon, v)$ 
such that there is at least one
pop transition of the form $\myep{{\it Z Y}}{Y'}$ or $Y = \Xfinal$.
Below we will prove that:
\begin{eqnarray}
\label{e:partialp}
p_{\myaut}(c) &=& 
\frac{ p_{\mygram}(\overline{v}) }{
	 \Sigma_{v'\in L_{X}}\ p_{\mygram}(\overline{v'}) }
\end{eqnarray}
Since a complete computation $c$ with output $v$ is of this form,
with $X = \Xinit$ and $Y= \Xfinal$, we
obtain the result we required to prove Theorem~\ref{t:sp},
where $D$ denotes the set of all 
complete derivations of CFG $\mygram$:
\begin{eqnarray} 
p_{\myaut}(c) &=& 
\frac{ p_{\mygram}(\overline{v}) }{ 
	\Sigma_{v'\in L_{\Xinit}}\ p_{\mygram}(\overline{v'}) } \\
&=& \frac{ p_{\mygram}(f(c)) }{
        \Sigma_{v'\in L_{\Xinit}}\ p_{\mygram}(f(v')) } \\
&=& \frac{ p_{\mygram}(f(c)) }{
        \Sigma_{d\in D}\ p_{\mygram}(d) } \\
&=&
p_{\mygram}(f(c))
\end{eqnarray}
We have used two properties of $f$ here. The first is that it
preserves the frequencies of symbols from $\myrule$, if considered as a
mapping from output strings to derivations.
The second property is that it can be considered as bijection from
complete computations to derivations. Lastly we have used
consistency of PCFG $(\mygram,  p_{\mygram})$, meaning that
$\Sigma_{d\in D}\ p_{\mygram}(d) = 1$.

For the proof of~(\ref{e:partialp}), we proceed by induction
on the length of $c$ and distinguish three cases.

Case~1: Consider a subcomputation $c$ consisting of zero transitions,
which naturally has output $v=\epsilon$,
with only configuration
$({\it X}, \epsilon, \epsilon)$, where there is at least one
pop transition of the form $\myep{{\it Z X}}{Z'}$ or $X = \Xfinal$.
We trivially have $p_{\myaut}(c)$ $=$ $1$
and
$\frac{ p_{\mygram}(\overline{v}) }{
         \Sigma_{v'\in L_{X}}\ p_{\mygram}(\overline{v'})}$ $=$
$\frac{ p_{\mygram}(\epsilon) }{
         \Sigma_{v'\in \{\epsilon\}}\ p_{\mygram}(\overline{v'})}$ $=$ $1$.

Case~2: Consider a subcomputation
$c=\tau_i c'$, where $({\it X}, x_iw, \epsilon)$
$\pdamove{\tau_i}$ $({\it Y_i}, w, y_i)$
$\pdamovesname{c'}$
$({\it Y'}, \epsilon, y_i v)$, such that there is at least one
pop transition of the form $\myep{{\it Z Y'}}{Z'}$ or $Y' = \Xfinal$.
The induction hypothesis states that:
\begin{eqnarray}
p_{\myaut}(c') &=&
\frac{ p_{\mygram}(\overline{v}) }{
         \Sigma_{v'\in L_{Y_i}}\ p_{\mygram}(\overline{v'}) }
\end{eqnarray}
If we combine this with the definition of $p_{\myaut}$, we obtain:
\begin{eqnarray} 
p_{\myaut}(c) &=& p_{\myaut}(\tau_i) \cdot  p_{\myaut}(c') \\
&=& 
\frac{  \Sigma_{v'\in L_{\tau_i}}\ p_{\mygram}(\overline{v'}) }{
        \Sigma_{v'\in L_{X}}\ p_{\mygram}(\overline{v'}) } \cdot
\frac{ p_{\mygram}(\overline{v}) }{
         \Sigma_{v'\in L_{Y_i}}\ p_{\mygram}(\overline{v'}) } \\
&=&
\frac{   p_{\mygram}(\overline{y_i}) \cdot \Sigma_{v'\in L_{Y_i}}\ 
			p_{\mygram}(\overline{v'}) }{
        \Sigma_{v'\in L_{X}}\ p_{\mygram}(\overline{v'}) } \cdot
\frac{ p_{\mygram}(\overline{v}) }{
         \Sigma_{v'\in L_{Y_i}}\ p_{\mygram}(\overline{v'}) } \\
&=&  
\frac{	p_{\mygram}(\overline{y_i}) \cdot p_{\mygram}(\overline{v}) }{
	 \Sigma_{v'\in L_{X}}\ p_{\mygram}(\overline{v'}) }  \\
&=& 
\frac{  p_{\mygram}(\overline{y_i v}) }{
	\Sigma_{v'\in L_{X}}\ p_{\mygram}(\overline{v'}) } 
\end{eqnarray}

\begin{figure}
\begin{center}
\epsfbox{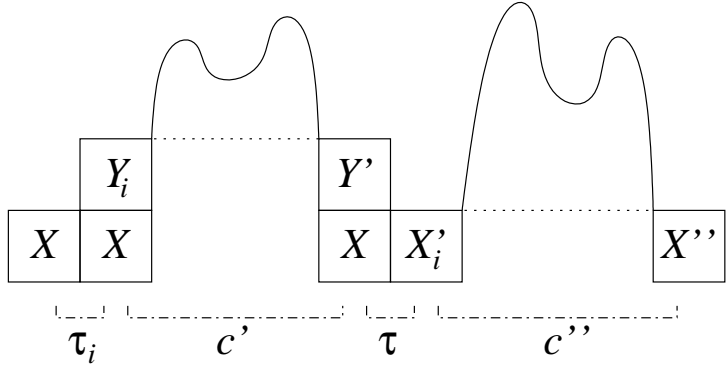}
\end{center}
\caption{Development of the stack in the computation
$c=\tau_i c' \tau c''$.}
\label{fig:stack}
\end{figure}

Case~3:  Consider a subcomputation
$c$ of the form $({\it X}, w, \epsilon)$ $\pdamove{\tau_i}$ 
$({\it XY_i}, w, \epsilon)$
$\pdamoves$
$({\it X''}, \epsilon, v)$ such that there is at least one
pop transition of the form $\myep{{\it Z X''}}{Z'}$ or $X'' = \Xfinal$.
Subcomputation $c$ can be decomposed in a unique way as 
$c=\tau_i c' \tau c''$,
consisting of an application of a push transition
$\tau_i = \myep{X}{X Y_i}$,
a subcomputation
$({\it Y_i}, w_1, \epsilon)$
$\pdamovesname{c'}$
$({\it Y'}, \epsilon, v_1)$,
an application of a pop transition
$\tau = \myep{XY'}{X_i'}$,
and a subcomputation 
$({\it X_i'}, w_2, \epsilon)$
$\pdamovesname{c''}$ 
$({\it X''}, \epsilon, v_2)$,
where $w=w_1w_2$ and $v=v_1 v_2$.
This is visualized in Figure~\ref{fig:stack}.

We can now use the induction hypothesis twice, resulting in:
\begin{eqnarray}
p_{\myaut}(c') &=&
\frac{ p_{\mygram}(\overline{v_1}) }{
         \Sigma_{v'_1\in L_{Y_i}}\ p_{\mygram}(\overline{v'_1}) }
\end{eqnarray}
and
\begin{eqnarray}
p_{\myaut}(c'') &=&
\frac{ p_{\mygram}(\overline{v_2}) }{
         \Sigma_{v'_2\in L_{X_i'}}\ p_{\mygram}(\overline{v'_2}) }
\end{eqnarray}

If we combine this with the definition of $p_{\myaut}$, 
we obtain:
\begin{eqnarray}
p_{\myaut}(c) &=& p_{\myaut}(\tau_i) \cdot p_{\myaut}(c') 
		\cdot p_{\myaut}(\tau)
		\cdot p_{\myaut}(c'') \\
&=& 
\frac{	\Sigma_{v'\in L_{\tau_i}}\ p_{\mygram}(\overline{v'}) }{
         \Sigma_{v'\in L_{X}}\ p_{\mygram}(\overline{v'}) }
\cdot 
\frac{  p_{\mygram}(\overline{v_1}) }{ 
         \Sigma_{v'_1\in L_{Y_i}}\ p_{\mygram}(\overline{v'_1}) } 
\cdot 1
\cdot  \frac{ p_{\mygram}(\overline{v_2}) }{
         \Sigma_{v'_2\in L_{X_i'}}\ p_{\mygram}(\overline{v'_2}) }
\end{eqnarray} 

Since $\myaut$ has the SPP, $X_i'$ is unique to $\tau_i$ and
the output strings in $L_{\tau_i}$ are precisely those that
can be obtained by concatenating
an output string in $L_{Y_i}$ and
an output string in $L_{X_i'}$.
Therefore $\Sigma_{v'\in L_{\tau_i}}\ p_{\mygram}(\overline{v'})$
$=$
$\Sigma_{v'_1\in L_{Y_i}} \Sigma_{v'_2\in L_{X_i'}}\ 
		p_{\mygram}(\overline{v'_1} \overline{v'_2})$
$=$
$\Sigma_{v'_1\in L_{Y_i}}\ p_{\mygram}(\overline{v'_1})$ $\cdot$
$\Sigma_{v'_2\in L_{X_i'}}\ p_{\mygram}(\overline{v'_2})$, 
and
\begin{eqnarray}
p_{\myaut}(c) &=&
\frac{  p_{\mygram}(\overline{v_1}) \cdot p_{\mygram}(\overline{v_2}) }{
	\Sigma_{v'\in L_{X}}\ p_{\mygram}(\overline{v'}) } \\
&=& 
\frac{ p_{\mygram}(\overline{v_1 v_2}) }{
	\Sigma_{v'\in L_{X}}\ p_{\mygram}(\overline{v'}) }  \\
&=& 
\frac{ p_{\mygram}(\overline{v}) }{
	\Sigma_{v'\in L_{X}}\ p_{\mygram}(\overline{v'}) } 
\end{eqnarray} 
This concludes the proof.~\closeproof

Note that the definition of $p_{\myaut}$ in the above proof relies on the
strings output by $\myaut$. This is the main reason
why we needed to consider push-down transducers rather
than push-down automata (defined below). 
Now assume an appropriate probability
function $p_{\myaut}$ has been found such that 
$(\myaut,p_{\myaut})$ is a PPDT that assigns the same
probabilities to computations as
the given PCFG assigns to the corresponding derivations, 
following the construction from the proof above. 
Then the probabilities assigned to strings over the input
alphabet are also equal.
We may subsequently ignore
the output strings if the application at hand merely requires
probabilistic recognition rather than probabilistic transduction,
or in other words, we may simplify push-down
transducers to push-down automata.

Formally, a {\em push-down automaton\/} (PDA) $\myaut$ is a 5-tuple
$(\myterm,$ $\mysym,$ $\Xinit,$ $\Xfinal,$ $\mytrans)$,
where $\myterm$ is the input alphabet, and
$\mysym,$ $\Xinit,$ $\Xfinal$ and $\mytrans$ are 
as in the definition of PDTs.
Push and pop transitions are as before, but swap transitions are 
simplified to the form
$\myscanrec{X}{x}{Y}$, where $x \in \{\epsilon\} \cup \Sigma$.
Computations are defined as in the case of PDTs, except that configurations
are now pairs $(\alpha,w)$ whereas they were triples $(\alpha,w,v)$
in the case of PDTs. A {\em probabilistic\/} push-down automaton (PPDA) is 
a pair $(\myaut,p_{\myaut})$, where $\myaut$ is a PDA
and $p_{\myaut}$ is a probability function
subject to the same constraints as in the case of
PPDTs.
Since the definitions of CPP and SPP for PDTs did not refer to output strings,
these notions carry over to PDAs in a straightforward way.

We define the size of a CFG as 
$\sum_{(A \de \alpha) \in \myrule} |A\alpha|$,
the total number of occurrences of
terminals and nonterminals in the set of rules.
Similarly,
we define the size of a PDA as 
$\sum_{(\myep{\alpha}{\beta})\in \mytrans} |\alpha\beta|+
\sum_{(\myscanrec{X}{x}{Y})\in \mytrans} |{\it XxY}|$,
the total number of occurrences of
stack symbols and terminals in the set of transitions.

Let $\myaut$ $=$ 
$(\myterm,$ $\mysym,$ $\Xinit,$ $\Xfinal,$ $\mytrans)$ be a 
PDA with both CPP and SPP.
We will now show that we can construct an equivalent CFG 
$\mygram$ = $(\myterm,$ $\mysym,$ $\Xinit,$ $\myrule)$
with size linear in the size of $\myaut$. 
The rules of this grammar are the following.
\begin{itemize}
\item
$X \de\it Y Z$
for each transition $\myep{X}{X Y}$, where
$Z$ is the unique stack symbol such that there is at
least one transition $\myep{X Y'}{Z}$ with $Y \pdagoto Y'$;
\item
$X \de x Y$
for each transition 
$\myscanrec{X}{x}{Y}$;
\item
$Y \de \epsilon$ for each stack symbol $Y$ 
such that there is at
least one transition $\myep{X Y}{Z}$ or such that $Y=\Xfinal$.
\end{itemize}
It is easy to see that there exists 
a bijection from complete computations of $\myaut$ to complete derivations
of $\mygram$, preserving the recognized/derived strings. 
Apart from an additional derivation step by rule
$\Xfinal \de \epsilon$, the complete derivations also have the
same length as the corresponding complete computations.

The above construction can straightforwardly be extended 
to probabilistic PDAs (PPDAs).
Let $(\myaut, p_{\myaut})$ be a PPDA with both CPP and SPP. 
Then we construct $\mygram$ as above, and further define
$p_{\mygram}$ such that
$p_{\mygram}(\pi) = p_{\myaut}(\tau)$ for rules
$\pi = X \de \it Y Z$ or $\pi = X \de x Y$ that we construct
out of transitions $\tau=\myep{X}{X Y}$ or $\tau=\myscanrec{X}{x}{Y}$,
respectively, in the first two items above.
We also define $p_{\mygram}(Y \de \epsilon) = 1$ 
for rules $Y \de \epsilon$ obtained in the third item above.
If $(\myaut, p_{\myaut})$ is reduced, proper and consistent then so is 
$(\mygram, p_{\mygram})$.

This leads to the observation that
parsing strategies with the
CPP and the SPP as well as
their probabilistic extensions can also be
described as grammar transformations, as follows. A given (P)CFG is mapped
to an equivalent (P)PDT by a (probabilistic)
parsing strategy. By ignoring the output components of
swap transitions we obtain a (P)PDA, which can be mapped to
an equivalent (P)CFG as shown above.
This observation gives rise to an extension with
probabilities of
the work on {\em covers\/} by \cite{NI80,LE89}.

It has been shown by \cite{GO82} that there is an infinite family of languages
with the following property. 
The sizes of the smallest CFGs generating those languages
are at least quadratically larger than the sizes of the smallest
equivalent PDAs. Note
that this increase in size cannot occur if PDAs satisfy 
both the CPP and the SPP, as we have shown above.

It is always possible to transform a PDA with the CPP but
without the SPP to an equivalent PDA with both CPP and SPP, 
by a construction that increases
the size of the PDA considerably (at least quadratically,
in the light of the above construction and \cite{GO82}).
However, such transformations in general
do not preserve parsing strategies and therefore are of minor interest
to the issues discussed in this paper.

The simple relationship between PDAs with both CPP and SPP 
on the one hand and CFGs on the other can 
be used to carry over algorithms originally 
designed for CFGs to PDAs or PDTs. One such application is the 
evaluation of the right-hand side of
equation~(\ref{e:normalized}) in the proof of
Theorem~\ref{t:sp}. Both the numerator and the denominator 
involve potentially infinite sets of subcomputations, and therefore
it is not immediately clear that the proof is constructive.
However, there are published algorithms to compute, for a
given PCFG $(\mygram',p_{\mygram'})$
that is not necessarily proper and a given
nonterminal $A$, the expression
$\Sigma_{w \in \Sigma^\ast}\ p_{\mygram'}(A \Rightarrow^\ast w)$,
or rather, to approximate it with arbitrary precision;
see \cite{BO73,ST95}. 
This can be used to compute e.g.\ 
$\Sigma_{v\in L_{X}}\ p_{\mygram}(\overline{v})$
in equation~(\ref{e:normalized}), as follows.

The first step 
is to map the PDT to a CFG $\mygram'$ as shown above.
We then define a function
$p_{\mygram'}$ that assigns probability
1 to all rules that we construct out of push and pop
transitions. We also let $p_{\mygram'}$ assign probability 
$p_{\mygram}(\overline{y})$ to a rule
$X \de x Y$ that we construct out of a scan transition
$\myscan{X}{x}{y}{Y}$. 
It is easy to see that, for any stack symbol $X$, we have
$\Sigma_{v\in L_{X}}\ p_{\mygram}(\overline{v}) =
\Sigma_{w \in \Sigma_1^\ast}\ p_{\mygram'}(X \Rightarrow^\ast w)$.
This allows our problem on the computations of probabilities
in the right-hand side of equation~(\ref{e:normalized})
to be reduced to a problem on PCFGs, which can be solved 
by existing algorithms as discussed above.  

\section{Parsing strategies with SPP}
\label{s:strong}

Many well-known parsing strategies with the CPP
also have the SPP,
such as top-down parsing \cite{HA78}, left-corner parsing \cite{RO70}
and PLR parsing \cite{SO79}, the first two of which we will
define explicitly here, whereas of the third we will merely 
present a sketch. A fourth strategy that we will discuss
is a combination of left-corner and top-down parsing, with special
computational properties.

In order to simplify the presentation, we allow a new type of
transition, without increasing the power of PDTs, viz.\
a combined push/swap transition of the form
$\myscan{X}{x}{y}{X Y}$. Such a transition can be seen as short-hand for
two transitions, the first of the form $\myep{X}{X Y_{x,y}}$,
where $Y_{x,y}$ is a new symbol not already in $\mysym$, and
the second of the form $\myscan{Y_{x,y}}{x}{y}{Y}$.

The first strategy we discuss is top-down parsing.
For a fixed CFG grammar $\mygram  = (\myterm,$ $\mynont,$ $S,$ $\myrule)$, 
we define $\mystrat_{\it TD}(\mygram) =
(\myaut,f)$. Here $\myaut$ $=$ $(\myterm,$ 
$\myrule,$ $\mysym,$ $[S \de\ \bul\sigma],$ $[S \de \sigma\bul],$ $\mytrans)$,
where $\mysym = \{ [A \de \alpha \bul \beta]\  |\ 
(A \de\alpha\beta) \in \myrule \}$;
these `dotted rules' are well-known from \cite{KN65,EA70}.
The transitions in $\mytrans$ are:
\begin{itemize}
\item $\myscan{[A \de \alpha \bul a \beta]}{a}{\epsilon}%
		{[A \de \alpha a \bul \beta]}$
for each rule $A \de \alpha a \beta$;
\item $\myscan{[A \de \alpha \bul B \beta]}{\epsilon}{\pi}%
		{[A \de \alpha \bul B \beta]\ [B\de\ \bul\gamma]}$
for each pair of rules 
$A \de \alpha B \beta$ and $\pi = B \de \gamma$;
\item $\myep{[A \de \alpha \bul B \beta]\ [B \de \gamma \bul]}%
		{[A \de \alpha B \bul \beta]}$.
\end{itemize}
The function $f$ is the identity function on strings over $\myrule$.
If seen as a function on computations,
then $f$ is a bijection from complete computations 
of $\myaut$ to complete derivations of $\mygram$, 
as required by the definition of `parsing strategy'. 

If $\mygram$ is reduced, then $\myaut$ clearly has the CPP.
That it also has the SPP can be
argued as follows. 
Let us first remark that if 
$[A \de \alpha \bul \beta]\pdagoto X$ 
for some stack symbols $[A \de \alpha \bul \beta]$ and $X$, 
then $X$ must be of the form
$[A \de \alpha \gamma\bul \delta]$, for some $\gamma$ and $\delta$ 
such that $\gamma\delta = \beta$.
Now, if there are three transitions
$\myep{X}{X Y}$, $\myep{X Y_1}{Z_1}$ and
$\myep{X Y_2}{Z_2}$ such that $Y\pdagoto Y_1$ and
$Y\pdagoto Y_2$, then 
$X$ must be of the form $[A \de \alpha \bul B \beta]$ 
and $Y$ of the form $[B\de\ \bul\gamma]$
(strictly speaking $[B\de\ \bul\gamma]_{\epsilon,\pi}$), 
$Y_1$ and $Y_2$ must both be $[B \de \gamma \bul]$,
and $Z_1$ and $Z_2$ must both be
$[A \de \alpha B \bul\beta]$.
Hence the SPP is satisfied.

Since $\mystrat_{\it TD}$ has both CPP and SPP, we
may apply Theorem~\ref{t:sp} to conclude that $\mystrat_{\it TD}$ can be 
extended to become a probabilistic parsing strategy. 
A direct construction of a top-down PPDT from a 
PCFG $(\mygram, p_{\mygram})$
is obtained by extending the above construction such that
probability 1 is assigned to all transitions produced by the first
and third items, and probability $p_{\mygram}(\pi)$ is assigned
to transitions produced by the second item.

The second strategy we discuss is left-corner (LC) parsing \cite{RO70}.
For a fixed CFG $\mygram= (\myterm,$ $\mynont,$ $S,$ $\myrule)$,
we define the binary relation $\LC$ over
$\myterm \cup \mynont$ by:
$X \LC A$ if and only if there is
an $\alpha \in (\myterm \cup \mynont)^\ast$
such that $(A \de X\alpha)\in\myrule$,
where $X \in \myterm \cup \mynont$.
We define the binary relation $\LCstar$ to be the reflexive and 
transitive closure of $\LC$. This implies that $a \LCstar a$ for all
$a \in \myterm$.

We now define $\mystrat_{\it LC}(\mygram) =
(\myaut,f)$. Here $\myaut$ $=$ $(\myterm,$ 
$\myrule\cup\{\dashv\},$ 
$\mysym,$ $[S \de\ \bul\sigma],$ $[S \de \sigma\bul],$ $\mytrans)$,
where $\mysym$ contains stack
symbols of the form $[A \de \alpha \bul \beta]$
where $(A \de\alpha\beta) \in \myrule$ such that 
$\alpha\neq\epsilon\vee A =S$,
and 
stack symbols of the form
$[A \de \alpha \bul Y\!\beta; X]$
where $(A \de\alpha Y\!\beta) \in \myrule$ and
$X,Y\in\myterm \cup \mynont$ such that $\alpha\neq\epsilon\vee A =S$
and $X \LCstar Y$.
The latter type of stack symbol indicates that
left corner $X$ of goal $Y$ in the right-hand side of rule
$A \de \alpha Y\! \beta$ has just been recognized.
The transitions in $\mytrans$ are:
\begin{itemize}
\item $\myscan{[A \de \alpha \bul Y\! \beta]}{a}{\epsilon}%
		{[A \de \alpha \bul Y\! \beta; a]}$
for each rule $A \de \alpha Y\! \beta$ and $a \in \myterm$
such that $\alpha\neq\epsilon\vee A=S$ and $a \LCstar Y$;
\item $\myscan{[A \de \alpha \bul B \beta]}{\epsilon}{\pi}%
		{[A \de \alpha \bul B \beta; C]}$
for each pair of rules $A \de \alpha B \beta$ and
$\pi = C \de \epsilon$ such that $\alpha\neq\epsilon\vee A=S$ and $C \LCstar B$;
\item $\myscan{[A \de \alpha \bul B \beta; X]}{\epsilon}{\pi}%
                {[A \de \alpha \bul B \beta; X]\ [C\de X \bul\gamma]}$
for each pair of rules 
$A \de \alpha B \beta$ and $\pi = C \de X \gamma$ such that 
$\alpha\neq\epsilon\vee A=S$ and $C \LCstar B$;
\item $\myep{[A \de \alpha \bul B \beta; X]\ [C\de X\gamma \bul]}%
		{[A \de \alpha \bul B \beta; C]}$
for each pair of rules
$A \de \alpha B \beta$ and $C \de X\gamma$ such that 
$\alpha\neq\epsilon\vee A=S$
and 
$C \LCstar B$;
\item $\myscan{[A \de \alpha \bul Y\! \beta; Y]}%
		{\epsilon}{\dashv}
                {[A \de \alpha Y \bul \beta]}$
for each rule $A \de \alpha Y\! \beta$ such that 
$\alpha\neq\epsilon\vee A=S$.
\end{itemize}
The function $f$ has to rearrange an output string to obtain
a complete derivation.
To make this possible, the output alphabet contains the
symbol $\dashv$ in addition to rules from $\myrule$. 
This symbol is used to mark the end of
an upward path of nodes in the parse tree 
each of which, except the last, is
the left-most daughter node of its mother node.
As explained in \cite{NI80}, in the absence of such a symbol,
it would be impossible to uniquely identify output strings with
derivations of the input.\footnote{%
In \cite[pp.~22--23]{NI80} a context-free grammar is considered
that consists of the set of rules 
$R = \{S \de {\it aS}, S \de {\it Sb}, S \de c\}$.
It is shown that any left-corner push-down 
transducer using only $R$ as output alphabet would
output at most one string for each input string, 
whereas there may be several 
derivations of the input, as the grammar is ambiguous.}

The function $f$ for the strategy $\mystrat_{\it LC}$
is defined by Figure~\ref{f:fLC}. Function $f$ is defined
in terms of function $f_{\it LC}$, which has two arguments.
The first argument, $d$, is either the empty string or
a subderivation that has already been constructed.
The second argument is a suffix of the output
string originally supplied as argument to $f$. 
Function $f_{\it LC}$ removes the first symbol $\pi$
from the output string, which will be
a rule $A \de X X_{1} \cdots X_l$ or $A \de \epsilon$.
In the former case, $d$ must be $\epsilon$ if $X\in \myterm_1$
and $d$ must be a subderivation from nonterminal $X$ otherwise.
The function is then called recursively zero or more times,
once for each nonterminal in $X_{1} \cdots X_l$,
to obtain more subderivations $d_i$, $1 \leq i \leq l$,
each of which is
obtained by consuming a subsequent part of the output string.
These subderivations are combined into a larger subderivation
$d' = \pi d d_{1} \cdots d_l$. Depending on the
question whether we encounter $\dashv$ as the immediately
following symbol of the output string, we return the
derivation $d'$ and the remainder $v'$ of the output string, or
call $\mystrat_{\it LC}$ recursively once more to
obtain a larger subderivation.
\begin{figure}[tp]
\begin{eqnarray*}
f(v) &=& d \\
&& {\rm where} \\
&& (d, \epsilon) = f_{\it LC}(\epsilon,v) \\
f_{\it LC}(d,\pi v_0) &=& (d'',v'') \\
&& {\rm where} \\
&& l \mbox{\ is such that\ }
        \pi = A \de X X_{1} \cdots X_l\ \mbox{\ or} \\
&& \hspace*{5ex} \pi = A \de \epsilon \wedge l = 0 \\
&& (d_{1}, v_{1}) =
                {\rm if\ } X_{1} \in \myterm_1
                {\rm \ then\ } (\epsilon, v_{0})
                {\rm \ else\ } f_{\it LC}(\epsilon, v_{0}) \\
&& \ldots \\
&& (d_{l}, v_{l}) =
                {\rm if\ } X_{l} \in \myterm_1
                {\rm \ then\ } (\epsilon, v_{l-1})
                {\rm \ else\ } f_{\it LC}(\epsilon, v_{l-1}) \\
&& d' = \pi d d_{1} \cdots d_l \\
&& (d'',v'') =
                {\rm if\ } {\dashv} v'  = v_{l}
                {\rm \ then\ } (d',v')
                {\rm \ else\ } f_{\it LC}(d',v_{l}) 
\end{eqnarray*}
\caption{Function $f$ for $\mystrat_{\it LC}$.}
\label{f:fLC}
\end{figure}

It can be easily shown that this strategy has the CPP.
Regarding the SPP, note that if there are two transitions
$\myscan{[A \de \alpha \bul B \beta; X]}{\epsilon}{\pi}%
                {[A \de \alpha \bul B \beta; X]\ [C\de X \bul\gamma]}$
and $\myep{[A \de \alpha \bul B \beta; X]\ Y_1}{Z_1}$ such that
$[C\de X \bul\gamma] \pdagoto Y_1$, then 
$Y_1$ must be $[C\de X \gamma\bul]$
and $Z_1$ must be $[A \de \alpha \bul B \beta; C]$, which means that
$Z_1$ is uniquely determined by the first transition.

Since $\mystrat_{\it LC}$ has both CPP and SPP, 
left-corner parsing can be extended to become a 
probabilistic parsing strategy. A direct construction of
probabilistic left-corner parsers from PCFGs has been presented
by \cite{TE95}.

Since at most two rules occur in each of the items above,
the size of a (probabilistic)
left-corner parser is $\order{|\mygram|^2}$, where
$|\mygram|$ denotes the size of $\mygram$.
This is the same complexity as that of the direct
construction by \cite{TE95}.
This is in contrast to a construction of `shift-reduce' PPDAs
out of PCFGs from \cite{AB99}, which were of size
$\order{|\mygram|^5}$.\footnote{This construction consisted
of a transformation to Chomsky normal form followed by 
a transformation to Greibach normal form (GNF) \cite{HA78}.
Its worse-case time complexity, established in
p.c.\ with David McAllester, is reached for a family
of CFGs $(\mygram_n)_{n \geq 2}$, defined by $\mygram_n =$
$(\{a_1,\ldots,a_n\},$ $\{A_1,\ldots,A_n\},$ $A_1,$ $\myrule)$,
where $\myrule$ contains the rules
$A_i \de A_{i+1}$, for $1 \leq i \leq n-1$,
$A_n \de A_1$,
and $A_i \de A_{i}\ A_{i}$ and
$A_i \de a_i$, for $1 \leq i \leq n$.
After transformation to GNF, the grammar
contains $n^5$ rules of the
form 
$A_{i_1}/A_{i_2} \de a_{i_3}\ A_{i_2}/A_{i_4}\ A_{i_1}/A_{i_5}$,
with $1 \leq i_1,i_2,i_3,i_4,i_5 \leq n$.
In \cite{BL99} a more economical transformation 
to Greibach normal form is given; straightforward
extension to probabilities leads to 
probabilistic parsers of the type considered by \cite{AB99} 
of size $\order{|\mygram|^4}$.}
The ``conjecture that
there is no {\em concise\/} translation of
PCFGs into shift-reduce PPDAs'' from \cite{AB99}
is made less significant by the earlier construction by \cite{TE95}
and our construction above.
It must be noted however that the `shift-reduce' model adhered to
by \cite{AB99} is more restrictive than the PDT models adhered
to by \cite{TE95} and by us.

When we look at upper bounds on the sizes of PPDAs (or PPDTs)
that describe the same probability distributations as given 
PCFGs, and compare these with the upper bounds for
(non-probabilistic) PDAs (or PDTs) for given CFGs,
we can make the following observation.
Theorem~\ref{t:cpp} states
that parsing strategies without the CPP cannot be extended to
become probabilistic. Furthermore, \cite{LE00} has shown that
for certain fixed languages the smallest
PDAs without the CPP are much smaller than
the smallest PDAs with the CPP. 
It may therefore appear that probabilistic PDAs
are in general larger than non-probabilistic ones.
However, the automata studied by
\cite{LE00} pertain to very specific languages, and at this
point there is little reason to believe that the demonstrated
results for these languages carry over to
any reasonable strategy for {\em general\/} CFGs.

The third parsing strategy that we discuss is PLR parsing \cite{SO79}.
Since it is very similar to LC parsing, 
we merely provide a sketch.
The stack symbols for PLR parsing are like those for LC parsing, 
except that the parts of rules following the dot are omitted.
Thus, instead of symbols of the form
$[A \de \alpha \bul \beta]$ and of the
form $[A \de \alpha \bul \beta; X]$, a PLR parser
manipulates stack symbols
$[A \de \alpha ]$ and $[A \de \alpha ; X]$, respectively.
That $\beta$ is omitted means that PLR parsers may postpone commitment
to one from two similar rules $A \de \alpha \beta$ and 
$A \de \alpha \beta'$ until the point is reached where $\beta$ and
$\beta'$ differ. In this sense PLR parsing
is less predictive than LC parsing,
although it still satisfies the 
strong predictiveness property, so that it can be extended to
become probabilistic.

There are two minor differences between the transitions of LC
parsers and those of PLR parsers. The first is the simplification of
stack symbols as explained above. The second is that for PLR, 
output of a rule is delayed until it is completely recognized.
The resulting output strings are right-most
derivations in reverse, which requires different functions $f$ than in
the case of LC parsing.
Note that right-most derivations can be effectively mapped
to corresponding parse trees, and parse trees can be effectively 
mapped to corresponding left-most derivations. 
Hence the required functions $f$ clearly exist.

The last strategy to be discussed in this section is a combination
of left-corner and top-down parsing. It has the special property
that, provided the fixed CFG is acyclic, 
the length of computations is bounded by a 
linear function on the length of the input, which
means that the parser cannot `loop' on any input.
Note that if the grammar is not acyclic, computations
of unbounded length cannot be avoided by any parsing strategy.
{}From this perspective, this parsing strategy, which we will
call {\em $\epsilon$-LC\/} parsing, is optimal.
It is based on \cite{NE93b}, and a
related idea for LR parsing was described by \cite{NE96e}.
The special termination properties of this strategy will be needed
in Section~\ref{s:prefix}.

We first define the binary relation $\LCep$ over
$\myterm \cup \mynont$ by:
$X \LCep A$ if and only if there are
$\alpha,\beta\in (\myterm \cup \mynont)^\ast$
such that $(A \de \alpha X\beta)\in\myrule$
and $\alpha \Rightarrow^\ast \epsilon$.
Relation $\LCep$ differs from the relation $\LC$ defined earlier
in that epsilon-generating
nonterminals at the beginning of a rule may be ignored.

The stack symbols are now of the form
$[A \de \alpha \bul \beta, \mu\bul\nu]$ or of the
form $[A \de \alpha \bul Y\! \beta, \mu\bul\nu; X]$.
Similar to the stack symbols for pure LC parsing, we
have $\alpha\neq\epsilon\vee A =S$
and $X \LCepstar Y$. Different is the additional dotted expression
$\mu\bul\nu$, which is such that $\mu\nu$ is
a string of epsilon-generating nonterminals, occurring at
the beginning of the right-hand side of a rule 
$A \de \mu\nu\alpha \beta$ or $A \de \mu\nu\alpha Y\!\beta$,
respectively.
The string $\mu\nu$ will be ignored in the
part of the strategy that behaves like left-corner parsing,
where $\mu=\epsilon$.
However, when the dot of the first dotted expression is at the end,
i.e., when we obtain a stack symbol of the form
$[A \de \alpha \bul, \bul\nu]$, then
top-down parsing will be activated to retrieve epsilon-generating
subderivations for the nonterminals in $\nu$, 
and the dot will move through $\nu$ from
left to right.\footnote{%
Although such subderivations can also be pre-compiled
during construction of the PDT,
we refrain from doing so since this could lead to
a PDT of exponential size.}

We have $\Xinit = [S \de\ \bul \sigma, \bul]$
and $\Xfinal = [S \de \sigma\bul, \bul]$, where for technical
reasons, and without loss of generality, we assume that
$\sigma$ does not contain any epsilon-generating nonterminals.
Next to the symbols from $\myrule$ and the symbol $\dashv$,
the output alphabet $\myterm_2$ also includes the set
of integers $\{0,\ldots,l-1\}$, where $l= |\alpha|$ for
a rule $(A \de \alpha) \in\myrule$ of maximal length;
the purpose of such integers will become clear below.
For the definition of the set of transitions, we will be less
precise than for $\mystrat_{\it TD}$ and 
$\mystrat_{\it LC}$, to prevent
cluttering up the presentation with details. 
We point out however that
in order to produce a reduced PDT from a reduced CFG, further side
conditions are needed for all items below:

\begin{itemize}
\item $\myscan{[A \de \alpha \bul Y\! \beta, \bul\mu]}{a}{\epsilon}%
                {[A \de \alpha \bul Y\! \beta, \bul\mu; a]}$
for $a \in \myterm$ such that $a \LCepstar Y$;
\item $\myscan{[A \de \alpha \bul B \beta, \bul\mu]}{\epsilon}{\pi 0}%
		{[A \de \alpha \bul B \beta, \bul\mu; C]}$
for $\pi = C \de \epsilon$ such that $C \LCstar B$;
\item $\myscan{[A \de \alpha \bul B \beta, \bul\mu; X]}{\epsilon}{\pi m}%
                {[A \de \alpha \bul B \beta, \bul\mu; X]\ 
		[C\de X \bul\gamma, \bul\mu']}$
for $\pi = C \de \mu' X \gamma$ such that
$C \LCepstar B$ and
$\mu'\Rightarrow^\ast\epsilon$, where $m= |\mu'|$;
\item $\myep{[A \de \alpha \bul B \beta, \bul\mu; X]\ 
		[C\de X\gamma \bul, \mu'\bul]}%
                {[A \de \alpha \bul B \beta, \bul\mu; C]}$;
\item $\myscan{[A \de \alpha \bul Y\! \beta, \bul\mu; Y]}%
                {\epsilon}{\dashv}
                {[A \de \alpha Y \bul \beta, \bul\mu]}$;
\item $\myscan{[A \de \alpha \bul, \mu \bul B \nu]}{\epsilon}{\pi}%
                {[A \de \alpha \bul, \mu \bul B \nu]\ 
		[B\de\ \bul, \bul\mu']}$
for $\pi = B \de \mu'$ such that $\mu'\Rightarrow^\ast\epsilon$;
\item $\myep{[A \de \alpha \bul, \mu \bul B \nu]\ [B\de\ \bul, \mu'\bul]}%
                {[A \de \alpha \bul, \mu B \bul \nu]}$.
\end{itemize}

The first five items are almost identical to the five
items we presented for $\mystrat_{\it LC}$,
except that strings $\mu$ of 
epsilon-generating nonterminals at the beginning of rules
are ignored. 
The length $m$ of a string $\mu$ is output just after 
the relevant grammar rule is output, in the second and third items.
This length $m$ will be needed to define function $f$ below.

The last two items follow a top-down strategy, but only for
epsilon-generating rules.
The produced transitions do what
was deferred by the left-corner part of the strategy:
they construct subderivations for the 
epsilon-generating nonterminals in strings $\mu$.

The function $f$, which produces a complete derivation
from an output string, is defined through two
auxiliary functions, viz.\
$\fepLC$ for the left-corner part and 
$\fepTD$ for the top-down
part, as shown in Figure~\ref{f:fepLC}.

\begin{figure}[tp]
\begin{eqnarray*}
f(v) &=& d \\
&& {\rm where} \\
&& (d, \epsilon) = \fepLC(\epsilon,v) \\
\fepLC(d,\pi m v_{0}) &=& (d'',v'') \\
&& {\rm where} \\
&& l \mbox{\ is such that\ } 
	\pi = A \de B_1 \cdots B_m X X_{1} \cdots X_l\ \mbox{\ or} \\
&& \hspace*{5ex} \pi = A \de \epsilon \wedge l = 0 \\
&& (d_{1}, v_{1}) = 
		{\rm if\ } X_{1} \in \myterm_1 
		{\rm \ then\ } (\epsilon, v_{0}) 
		{\rm \ else\ } \fepLC(\epsilon, v_{0}) \\
&& \ldots \\
&& (d_{l}, v_{l}) =
                {\rm if\ } X_{l} \in \myterm_1 
                {\rm \ then\ } (\epsilon, v_{l-1}) 
                {\rm \ else\ } \fepLC(\epsilon, v_{l-1}) \\
&& (d'_1,v_{l+1}) = \fepTD(v_{l}) \\
&& \ldots \\
&& (d'_m,v_{l+m}) = \fepTD(v_{l+m-1}) \\
&& d' = \pi d'_1 \cdots d'_m d d_{1} \cdots d_l \\
&& (d'',v'') =
		{\rm if\ } {\dashv} v'  = v_{l+m}
		{\rm \ then\ } (d',v') 
		{\rm \ else\ } \fepLC(d',v_{l+m}) \\
\fepTD(v) &=& (\pi d_1 \cdots d_l,v_l) \\
&& {\rm where} \\
&& \pi v_{0} = v \\
&& l \mbox{\ is such that\ } \pi = A \de B_{1} \cdots B_l  \\
&& (d_1,v_1) = \fepTD(v_0) \\
&& \ldots \\
&& (d_l,v_l) = \fepTD(v_{l-1})
\end{eqnarray*}

\caption{Function $f$ for $\stratepLC$.}
\label{f:fepLC}
\end{figure}

The function $\fepLC$ is similar to $f_{\it LC}$ defined in 
Figure~\ref{f:fLC}. The main difference is that now 
subderivations deriving $\epsilon$
for the first $m$ nonterminals in the right-hand side
of a rule are obtained by calls of the function $\fepTD$.
For a suffix $v$ of an output string,
$\fepTD(v)$ yields a pair $(\pi d_1 \cdots d_l,v_l)$
such that $v= \pi d_1 d_2 \cdots d_lv_l$. In other words,
$\fepTD$ does nothing more than split its argument into
two parts. The length of the first part $\pi d_1 \cdots d_l$
depends on the
length $l$ of the right-hand side of rule $\pi$ and
on the lengths of right-hand sides of rules that
are visited recursively.

It can be easily seen that $\stratepLC$
has both CPP and SPP. The size of a produced
PDT is now $\order{|\mygram|^3}$, rather than
$\order{|\mygram|^2}$ as in the case of $\mystrat_{\it LC}$.

\section{Parsing strategies without SPP}
\label{s:nonstrong}

In this section
we show that the absence of the strong predictiveness property
may mean that a parsing strategy with the CPP
cannot be extended to become a
probabilistic parsing strategy. We first illustrate this for
LR(0) parsing, formalized as a
parsing strategy $\mystrat_{\it LR}$,
which has the CPP but not the SPP, 
as we will see.
We assume the reader is familiar with LR parsing; see \cite{SI90}.

We take a PCFG $(\mygram, p_{\mygram})$ defined by:
$$
\begin{array}{c@{\;=\;}ll}
\pi_{S} & S \de {\it AB}, & p_{\mygram}(\pi_{S}) = 1 \\[.1ex]
\pi_{A_1} & A \de {\it aC}, & p_{\mygram}(\pi_{A_1}) = \frac{1}{3} \\[.1ex]
\pi_{A_2} & A \de {\it aD}, & p_{\mygram}(\pi_{A_2}) = \frac{2}{3} \\[.1ex]
\pi_{B_1} & B \de {\it bC}, & p_{\mygram}(\pi_{B_1}) = \frac{2}{3} \\[.1ex]
\pi_{B_2} & B \de {\it bD}, & p_{\mygram}(\pi_{B_2}) = \frac{1}{3} \\[.1ex]
\pi_{C} & C \de {\it xc}, & p_{\mygram}(\pi_{C}) = 1 \\[.1ex]
\pi_{D} & D \de {\it xd}, & p_{\mygram}(\pi_{D}) = 1
\end{array}
$$
Note that this grammar generates a finite language.

We will not present the entire LR automaton $\myaut$, 
with $\mystrat_{\it LR}(\mygram) = (\myaut,f)$ for some $f$,
but we merely mention two of its key transitions, which
represent shift actions over $c$ and $d$:
$$
\begin{array}{c@{\;=\;}l}
\tau_{c} & \myscan{\{C\de x\bul c, D\de x\bul d\}}%
		{c}{\epsilon}%
		{\{C\de x\bul c, D\de x\bul d\}\ \{C \de xc\bul\}} \\
\tau_{d} & \myscan{\{C\de x\bul c, D\de x\bul d\}}%
                {d}{\epsilon}%
                {\{C\de x\bul c, D\de x\bul d\}\ \{D \de xd\bul\}}
\end{array}
$$
(We denote LR states by their sets of kernel items, as usual.)

Take a probability function $p_{\myaut}$
such that $(\myaut, p_{\myaut})$ is a proper PPDT.
It can be easily seen 
that $p_{\myaut}$ must assign 1 to all
transitions except $\tau_{c}$ and $\tau_{d}$, since that is the only
pair of distinct transitions that can be applied for one and the
same top-of-stack symbol,
viz.\ $\{C\de x\bul c,D\de x\bul d\}$.
 
However,
$\frac{p_{\mygram}({\it axcbxd})}{p_{\mygram}({\it axdbxc})} =
\frac{p_{\mygram}(\pi_{A_1}) \cdot  p_{\mygram}(\pi_{B_2})}%
{p_{\mygram}(\pi_{A_2}) \cdot  p_{\mygram}(\pi_{B_1})} =
\frac{\frac{1}{3}\cdot\frac{1}{3}}{\frac{2}{3}\cdot\frac{2}{3}} = 
\frac{1}{4}$
but
$\frac{p_{\myaut}({\it axcbxd})}{p_{\myaut}({\it axdbxc})} =
\frac{p_{\myaut}(\tau_{c}) \cdot  p_{\myaut}(\tau_{d})}%
{p_{\myaut}(\tau_{d}) \cdot  p_{\myaut}(\tau_{c})} = 1 \neq \frac{1}{4}$.
This shows that there is no $p_{\myaut}$ such that
$(\myaut, p_{\myaut})$ assigns the same
probabilities to strings over $\myterm$ as $(\mygram, p_{\mygram})$.
It follows that
the LR strategy cannot be extended to become a probabilistic
parsing strategy.

Note that for $\mygram$ as above, $p_{\mygram}(\pi_{A_1})$
and $p_{\mygram}(\pi_{B_1})$ can be freely chosen, and this
choice determines the other values of $p_{\mygram}$, so we have
two free parameters. For $\myaut$ however, there is only one
free parameter in the choice of $p_{\myaut}$.
This is in conflict with an underlying assumption of existing work
on probabilistic LR parsing, by e.g.\ \cite{BR93} and \cite{IN00},
viz.\ that LR parsers would allow more fine-grained probability
distributions than CFGs. However, for some practical grammars 
from the area of natural language processing,
\cite{SO99} has shown that LR parsers do allow
more accurate probability distributions than the CFGs from which
they were constructed, if probability functions are estimated from
corpora.

By way of Theorem~\ref{t:sp}, it follows indirectly from
the above that LR parsing lacks the SPP. 
For the somewhat simpler ELR parsing strategy,
to be discussed next,
we will give a direct explanation of why it lacks the SPP.
A direct explanation for LR parsing is much more involved and
therefore is not reported here, although the argument is essentially
of the same nature as the one we discuss for ELR parsing.

The ELR parsing strategy is not as well-known as LR parsing.
It was originally 
formulated as a parsing strategy for extended CFGs \cite{PU81,LE89},
but its restriction to normal CFGs is interesting in its
own right, as argued by \cite{NE94a}. 
ELR parsing for CFGs is also related to the tabular algorithm
from \cite{VO88}.

Concerning the representation of right-hand sides of rules,
stack symbols
for ELR parsing are similar to those for PLR parsing:
only the part of a right-hand side is represented
that consists of the grammar symbols that have been processed.
Different from LC and PLR parsing is however that a 
stack symbol for ELR parsing contains
a set consisting of one or more nonterminals from
the left-hand sides of pairwise similar rules, 
rather than a single such nonterminal. 
This allows the commitment to certain rules,
and in particular to their left-hand sides, to be
postponed even longer than for LC and PLR parsing.

Thus, for a given CFG $\mygram  = (\myterm,$ $\mynont,$ $S,$ $\myrule)$,
we construct a pair $\mystrat_{\it ELR}(\mygram) =
(\myaut,f)$. Here $\myaut$ $=$ $(\myterm,$
$\myrule,$ $\mysym,$ $[\{S\} \de\epsilon],$ $[\{S\} \de \sigma],$ $\mytrans)$,
where $\mysym$ is a subset of $\{ [\mynontset \de \alpha ]\  |\
\mynontset \subseteq \mynont \wedge 
\forall A \in \mynontset \exists \beta[(A \de\alpha\beta) \in \myrule] \}$ $\cup$
$\{ [\mynontset \de \alpha; B]\  |\
\mynontset \subseteq\nobreak \mynont \wedge
\forall A \in\nobreak\mynontset 
\exists\beta[(A \de\alpha\beta) \in \myrule
\wedge B \in \mynont] \}$.

We provide simultaneous inductive definitions of $\mysym$ and
$\mytrans$:
\begin{itemize}
\item $[\{S\} \de\epsilon]\in \mysym$;
\item For $[\mynontset \de \alpha ] \in \mysym$, 
rule $A \de \alpha Y \beta$ and $a\in\myterm$ such that 
$A \in \mynontset$ and $a \LCstar Y$, let
$[\mynontset \de \alpha; a] \in \mysym$ and
$\myscan{[\mynontset \de \alpha ]}{a}{\epsilon}%
                {[\mynontset \de \alpha; a]} \in \mytrans$;
\item For $[\mynontset \de \alpha] \in \mysym$,
rules $A \de \alpha B \beta$ and
$\pi = C \de \epsilon$ such that $A\in\mynontset$ and
$C \LCstar B$, let
$[\mynontset \de \alpha ; C] \in \mysym$ and 
$\myscan{[\mynontset \de \alpha ]}{\epsilon}{\pi}%
                {[\mynontset \de \alpha ; C]} \in \mytrans$;
\item For $[\mynontset_1 \de \alpha ; X] \in \mysym$ and 
$\mynontset_2 = \{ C\ |\ 
\exists (A \de \alpha B \beta)\in\myrule[A\in \mynontset_1 \wedge
C \de X \gamma \wedge C\LCstar B] \} \neq \emptyset$, let
$[\mynontset_2\de X ]\in \mysym$ and 
$\myep{[\mynontset_1 \de \alpha ; X]}%
                {[\mynontset_1 \de \alpha;X]\ [\mynontset_2\de X ]} \in \mytrans$;
\item For $[\mynontset_1 \de \alpha;X], [\mynontset_2\de X\gamma ] \in \mysym$,
rules $A \de \alpha B \beta$ and $\pi = C \de X\gamma$ such that
$A \in \mynontset_1$, $C \in \mynontset_2$ and $C \LCstar B$, let
$[\mynontset_1 \de \alpha; C] \in \mysym$ and 
$\myscan{[\mynontset_1 \de \alpha;X]\ [\mynontset_2\de X\gamma ]}{\epsilon}{\pi}
                {[\mynontset_1 \de \alpha; C]} \in \mytrans$;
\item For $[\mynontset_1 \de \alpha ; Y] \in \mysym$ and 
$\mynontset_2 = \{ A\in\mynontset_1\ |\ 
\exists \beta[(A \de \alpha Y \beta)\in\myrule] \}
\neq \emptyset$, let
$[\mynontset_2 \de \alpha Y ] \in \mysym$ and
$\myep{[\mynontset_1 \de \alpha ; Y]}%
                {[\mynontset_2 \de \alpha Y ]} \in \mytrans$.
\end{itemize}
Note that the last five items are very similar to the five items 
for LC parsing. In the second last item, we have assumed
the availability of combined pop/swap transitions of the form
$\myscan{X Y}{x}{y}{Z}$. Such a transition can be seen as short-hand for
two transitions, the first of the form $\myep{X Y}{Z_{x,y}}$,
where $Z_{x,y}$ is a new symbol not already in $\mysym$, and
the second of the form $\myscan{Z_{x,y}}{x}{y}{Z}$.

The function $f$ is defined as in the case of PLR parsing, and
turns a complete right-most derivation in
reverse into a complete derivation.

ELR parsing has the CPP but, like LR parsing,
it lacks the SPP. The problem is caused by
transitions of the form
$\myscan{[\mynontset_1 \de \alpha;X]\ [\mynontset_2\de X\gamma ]}{\epsilon}{\pi}
                {[\mynontset_1 \de \alpha; C]}$.
Intuitively, a subcomputation that recognizes $\gamma$,
directly after recognition of $X$, only commits to
a choice of the left-hand side nonterminal $C$ from
$\mynontset_2$ after $\gamma$ has been
completely recognized, and this choice is communicated
to lower areas of the stack through this pop transition.

\begin{figure}[tp]
$$
\begin{array}{cl}
\comment{\tau_{a}} & \myscan{[\{S\}\de\epsilon]}{a}{\epsilon}{[\{S\}\de\epsilon;a]} \\
\comment{\tau_{A}} & \myep{[\{S\}\de\epsilon;a]}{[\{S\}\de\epsilon;a]\ [\{A\}\de a]} \\
\comment{\tau_{A/x}} & \myscan{[\{A\}\de a]}{x}{\epsilon}{[\{A\}\de a; x]} \\
\comment{\tau'_{A/x}} & \myep{[\{A\}\de a;x]}{[\{A\}\de a;x]\ [\{C,D\}\de x]} \\
\tau_{c}\ = & \myscan{[\{C,D\}\de x]}{c}{\epsilon}{[\{C,D\}\de x}; c] \\
\tau_{d}\ = & \myscan{[\{C,D\}\de x]}{d}{\epsilon}{[\{C,D\}\de x}; d] \\
\comment{\tau_{C}} & \myep{[\{C,D\}\de x;c]}{[\{C\}\de xc]} \\
\comment{\tau_{A/C}} &  \myscan{[\{A\}\de a;x]\ [\{C\}\de xc]}%
		{\epsilon}{\pi_{C}}{[\{A\}\de a; C]} \\
\comment{\tau'_{A/C}} & \myep{[\{A\}\de a; C]}[\{A\}\de a C] \\
\comment{\tau_{A_1}} & \myscan{[\{S\}\de\epsilon;a]\ [\{A\}\de aC]}%
		{\epsilon}{\pi_{A_1}}{[\{S\}\de \epsilon; A]} \\
\comment{\tau_{D}} & \myep{[\{C,D\}\de x;d]}{[\{D\}\de xd]}  \\
\comment{\tau_{A/D}} &  \myscan{[\{A\}\de a;x]\ [\{D\}\de xd]}%
		{\epsilon}{\pi_{D}}{[\{A\}\de a; D]} \\
\comment{\tau'_{A/D}} & \myep{[\{A\}\de a; D]}[\{A\}\de a D] \\
\comment{\tau_{A_2}} & \myscan{[\{S\}\de\epsilon;a]\ [\{A\}\de aD]}%
		{\epsilon}{\pi_{A_2}}{[\{S\}\de \epsilon; A]} \\
\comment{\tau_{S/A}} & \myep{[\{S\}\de \epsilon; A]}{[\{S\}\de A]} \\
\comment{\tau_{b}} & \myscan{[\{S\}\de A]}{b}{\epsilon}{[\{S\}\de A; b]} \\
\comment{\tau_{B}} & \myep{[\{S\}\de A;b]}{[\{S\}\de A;b]\ [\{B\}\de b]} \\
\comment{\tau_{B/x}} & \myscan{[\{B\}\de b]}{x}{\epsilon}{[\{B\}\de b; x]} \\
\comment{\tau'_{B/x}} & \myep{[\{B\}\de b;x]}{[\{B\}\de b;x]\ [\{C,D\}\de x]} \\
\comment{\tau_{B/C}} &  \myscan{[\{B\}\de b;x]\ [\{C\}\de xc]}%
		{\epsilon}{\pi_{C}}{[\{B\}\de b; C]} \\
\comment{\tau'_{B/C}} & \myep{[\{B\}\de b; C]}[\{B\}\de b C] \\
\comment{\tau_{B_1}} & \myscan{[\{S\}\de A;b]\ [\{B\}\de bC]}%
		{\epsilon}{\pi_{B_1}}{[\{S\}\de A;B]} \\
\comment{\tau_{B/D}} &  \myscan{[\{B\}\de b;x]\ [\{D\}\de xd]}%
		{\epsilon}{\pi_{D}}{[\{B\}\de b; D]} \\
\comment{\tau'_{B/D}} & \myep{[\{B\}\de b; D]}[\{B\}\de b D] \\
\comment{\tau_{B_2}} & \myscan{[\{S\}\de A;b]\ [\{B\}\de bD]}%
		{\epsilon}{\pi_{B_2}}{[\{S\}\de A;B]} \\
\comment{\tau_{S/B}} & \myep{[\{S\}\de A; B]}{[\{S\}\de AB]} \\
\end{array}
$$
\caption{Transitions for ELR parsing strategy.}
\label{f:ELRtrans}
\end{figure}

That ELR parsing can indeed not be extended to a probabilistic
parsing strategy can be shown by considering the same
CFG as above. From the set of transitions, shown in 
Figure~\ref{f:ELRtrans},
we restrict our attention to the following two:
$$
\begin{array}{c@{\;=\;}l}
\tau_{c} & \myscan{[\{C,D\}\de x]}{c}{\epsilon}{[\{C,D\}\de x}; c] \\
\tau_{d} & \myscan{[\{C,D\}\de x]}{d}{\epsilon}{[\{C,D\}\de x}; d] 
\end{array}
$$
This is the only pair of transitions that can be applied for
one and the same top-of-stack.
The rest of the proof is identical to that in the case of
LR parsing.

Problems with the extension of ELR parsing to become
a probabilistic parsing
strategy have been pointed out before by \cite{TE97},
who furthermore proposed an alternative type of probabilistic
push-down automaton that is capable of computing multiple 
probabilities for each subderivation. 
However, since a transition of such an automaton may perform an
unbounded number of elementary computations on probabilities, we 
feel this automaton model cannot realistically express
the behaviour of probabilistic parsers,
and therefore it will not be considered further here.

\section{Extension in the wide sense}
\label{s:wide}

The main result from the previous section is that,
in general,
there is no construction of probabilistic LR parsers 
from PCFGs such that, 
firstly, a probabilistic LR parser has the same set of
transitions as the LR parser that would be constructed from the CFG in
the non-probabilistic case and,
secondly, the probabilistic LR parser
has the same probability distribution as the given PCFG.

There is a construction proposed by \cite{WR91,WR91a,NG91}
that operates under different assumptions. In particular, a
probabilistic LR parser constructed from a certain PCFG
may possess several `copies' of one and the same 
LR state from the (non-probabilistic) LR parser constructed from
the CFG, 
each annotated with some additional information to
distinguish it from other copies of the same LR state. 
Each such copy behaves as the corresponding LR state from the
LR parser if we neglect probabilities. 
Transitions may however
obtain different probabilities if they operate on different copies
of identical LR states, based on the additional information
attached to the LR states.

By this construction,
there are many PCFGs for which one may obtain a
probabilistic LR parser that describes the same
probability distribution. This even holds
for the PCFG we discussed in the previous section, although
we have shown that a probabilistic LR parser {\em without\/}
an extended LR state set could not describe the same
probability distribution.
A serious problem with this approach is however that 
the required number of copies of each LR state is potentially infinite.

In this section we formulate these observations in terms of
general parsing strategies and a wider notion of
extension to probabilistic parsing strategies. We also
show that the above-mentioned problem with
infinite numbers of states is inherent in LR parsing, rather
than due to the particular construction of LR parsers from
PCFGs by \cite{WR91,WR91a,NG91}.

We first introduce some auxiliary notation and terminology. 
Let $\myaut$ and $\myaut'$ be two PDTs and
let $g$ be a function mapping
the stack symbols of $\myaut'$
to the stack symbols of $\myaut$.
If $\tau$ is a transition of the form $\myep{X}{X Y}$,
$\myep{\it Y X}{Z}$ or $\myscan{X}{x}{y}{Y}$ from $\myaut'$,
then we let $g(\tau)$ denote a transition of the form
$\myep{g(X)}{g(X) g(Y)}$,
$\myep{\it g(Y) g(X)}{g(Z)}$ or $\myscan{g(X)}{x}{y}{g(Y)}$, respectively.
This effectively extends $g$ to a function from transitions to
transitions. 
Note that a transition $g(\tau)$ may, but need not be a
transition from $\myaut$.
In the same vein, we extend $g$ to
a function from computations of $\myaut'$ to
sequences of transitions (which may, but need not be
computations of $\myaut$),
by applying $g$ element-wise as a function on transitions.

For PDTs $\myaut$ $=$
$(\myterm_1,$ $\myterm_2,$ $\mysym,$ $\Xinit,$ $\Xfinal,$ $\mytrans)$
and $\myaut'$ $=$
$(\myterm'_1,$ $\myterm'_2,$ $\mysym',$ $\Xinit',$ $\Xfinal',$ $\mytrans')$,
we say
$\myaut'$
is an {\em expansion\/} of $\myaut$
if $\myterm'_1=\myterm_1$, $\myterm'_2 = \myterm_2$ and there is a function
$g$ such that:
\begin{itemize}
\item $g$ is a surjective function from $\mysym'$ to $\mysym$.
\item Extended to transitions, 
$g$ is a surjective function from $\mytrans'$ to $\mytrans$.
\item Extended to computations,
$g$ is a bijective function from the set of computations of
$\myaut'$ to the set of computations of $\myaut$.
\end{itemize}
In other words, for each stack symbol from $\mysym$, 
$\mysym'$ may contain one or more corresponding
stack symbols. The language that
is accepted and the output strings that are produced for given input
strings remain the same however. Furthermore, that $g$ is a bijection 
on computations implies that the behaviour of the two
automata is identical in terms of e.g.\ the length of
computations and the amount of nondeterminism encountered within
those computations.

To illustrate these definitions, assume we have an arbitrary
PDT $\myaut$. We construct a second PDT $\myaut'$ that is an
expansion of $\myaut$. It has the
same input and output alphabets, and for each stack symbol
$X$ from $\myaut$, $\myaut'$ has two stack symbols $(X,0)$ and
$(X,1)$. A second component $0$ signifies that the distance 
of the stack symbol to the bottom of the
stack is even, and $1$ that it is odd.
Naturally, if $\Xinit$ and $\Xfinal$ are the initial and final stack symbols
of $\myaut$, we choose the initial and final stack symbols of $\myaut'$ to be
$(\Xinit,0)$ and $(\Xfinal,0)$, as they have distance 0 to the
bottom of the stack.
For each transition of the form $\myep{X}{X Y}$, 
$\myep{\it Y X}{Z}$ or $\myscan{X}{x}{y}{Y}$ from $\myaut$,
we let $\myaut'$ have the transitions
$\myep{(X,i)}{(X,i) (Y,1-i)}$,
$\myep{(Y,i) (X,1-i)}{(Z,i)}$ or $\myscan{(X,i)}{x}{y}{(Y,i)}$, 
respectively, for both $i=0$ and $i=1$. 
Obviously, the function $g$ mapping stack symbols
from $\myaut'$ to stack symbols from $\myaut$ is given
by $g((X,i))=X$ for all $X$ and $i\in\{0,1\}$.

We now come to the central definition of this section.
We say that probabilistic parsing strategy $\mystrat'$
is an {\em extension in the wide sense\/} of parsing strategy 
$\mystrat$ if for each reduced CFG $\mygram$ and 
probability function $p_{\mygram}$ we have
$\mystrat(\mygram)=(\myaut, f)$ if and only if
$\mystrat'(\mygram, p_{\mygram})=(\myaut', p_{\myaut'}, f)$
for some $\myaut'$ that is an expansion of $\myaut$
and some $p_{\myaut'}$. This definition allows more 
probabilistic parsing strategies $\mystrat'$ to be related to a given
strategy $\mystrat$ than the definition of extension from 
Section~\ref{s:strategy}.

LR parsing however, which we know can not be extended to a 
probabilistic strategy in the narrow sense from Section~\ref{s:strategy}, 
can neither be
extended in the wide sense to a probabilistic parsing strategy.
To prove this,
consider the following PCFG $(\mygram,p_{\mygram})$, 
taken from \cite{WR91} with minor modifications:
$$
\begin{array}{c@{\;=\;}ll}
\pi_{S} & S \de A, & p_{\mygram}(\pi_{S}) = 1 \\[.1ex]
\pi_{A_1} & A \de B, & p_{\mygram}(\pi_{A_1}) = \frac{1}{2} \\[.1ex]
\pi_{A_2} & A \de C, & p_{\mygram}(\pi_{A_2}) = \frac{1}{2} \\[.1ex]
\pi_{B_1} & B \de {\it aB}, & p_{\mygram}(\pi_{B_1}) = \frac{1}{3} \\[.1ex]
\pi_{B_2} & B \de {\it b}, & p_{\mygram}(\pi_{B_2}) = \frac{2}{3} \\[.1ex]
\pi_{C_1} & C \de {\it aC}, & p_{\mygram}(\pi_{C_1}) = \frac{2}{3} \\[.1ex]
\pi_{C_2} & C \de {\it c}, & p_{\mygram}(\pi_{C_2}) = \frac{1}{3}
\end{array}
$$
The CFG $\mygram$ generates strings of the form $a^n b$ and $a^n c$ for
any $n \geq 0$. Observe that
$\frac{p_{\mygram}(a^n b)}{p_{\mygram}(a^n c)}$ $=$
$\frac{ \frac{1}{2} \cdot
		\left(\frac{1}{3}\right)^{n} \cdot \frac{2}{3} }{
	\frac{1}{2} \cdot
                \left(\frac{2}{3}\right)^{n} \cdot \frac{1}{3} }$ $=$
$\left( \frac{1}{2} \right)^{n-1}$. 

Let $\myaut$ be such that $\mystrat_{\it LR}(\mygram)= (\myaut,f)$ and
consider input strings of the form $a^n b$ and $a^n c$, $n \geq 1$. 
After scanning the first $n$ symbols, $\myaut$
reaches a configuration where the top-of-stack $X$ is
given by the set of (kernel) items:
$$
X=\{ B \de a \bul B, C \de a \bul C \}
$$

There are three applicable transitions, representing shift
actions over $a$, $b$ and $c$, given by:
$$
\begin{array}{c@{\;=\;}l}
\tau_a & \myscan{X}{a}{\epsilon}{X\ X} \\
\tau_b & \myscan{X}{b}{\epsilon}{X\ \{B \de b\bul\}} \\
\tau_c & \myscan{X}{c}{\epsilon}{X\ \{C \de c\bul\}} 
\end{array}
$$
After reading $b$ or $c$,
the remaining transitions are fully deterministic.

For a PDT $\myaut'$ that is an expansion of $\myaut$, we may have
different stack symbols that are all mapped to $X$ by function $g$. 
These stack symbols can be referred to as 
$X_n$, which occur as top-of-stack
after scanning the first $n$ symbols of $a^n b$ or $a^n c$, $n \geq 1$.
We refer to the applicable transitions with top-of-stack $X_n$ as:
$$
\begin{array}{c@{\;=\;}l}
\tau_{a,n} & \myscan{X_n}{a}{\epsilon}{X_n\ X_{n+1}} \\
\tau_{b,n} & \myscan{X_n}{b}{\epsilon}{X_n\ \{B \de b\bul\}_n} \\
\tau_{c,n} & \myscan{X_n}{c}{\epsilon}{X_n\ \{C \de c\bul\}_n}
\end{array}
$$
for certain stack symbols $\{B \de b\bul\}_n$ and
$\{C \de c\bul\}_n$ that $g$ maps to $\{B \de b\bul\}$ and
$\{C \de c\bul\}$, respectively.

Now let us assume we have a probability function $p_{\myaut'}$
such that $(\myaut',p_{\myaut'})$ is a PPDT.
Since the application of either $\tau_{b,n}$ or $\tau_{c,n}$ is
the only nondeterministic step 
that distinguishes recognition of $a^n b$ from
recognition of $a^n c$, $n \geq 1$, it follows that
$\frac{p_{\myaut}(a^n b)}{p_{\myaut}(a^n c)}$ $=$
$\frac{p_{\myaut}(\tau_{b,n})}{p_{\myaut}(\tau_{c,n})}$.
If $(\myaut',p_{\myaut'})$ assigns the same probabilities
to strings over alphabet $\{a,b,c\}$ as $(\mygram,p_{\mygram})$,
then $\frac{p_{\myaut}(\tau_{b,n})}{p_{\myaut}(\tau_{c,n})}$
must be equal to $\frac{p_{\mygram}(a^n b)}{p_{\mygram}(a^n c)}$ $=$
$\left( \frac{1}{2} \right)^{n-1}$ for each 
$n\geq 1$. Since $\left( \frac{1}{2} \right)^{n-1}$ is a different
value for each $n$ however, this would require $\myaut'$ to possess
infinitely many stack symbols, which is in conflict with the definition
of push-down transducers.

This shows that no probability function $p_{\myaut'}$ exists
for any expansion $\myaut'$ of $\myaut$ such that
$(\myaut',p_{\myaut'})$ assigns the same probabilities
to strings over the alphabet as $(\mygram,p_{\mygram})$,
and therefore LR parsing cannot be extended in the wide sense to
become a probabilistic parsing strategy. With only minor changes
to the proof, the same can be shown for ELR parsing.

\section{Prefix probabilities}
\label{s:prefix}

In this section we show that the behaviour of PPDTs on input
can be simulated by dynamic programming. 
We also show how dynamic programming can be used for
computing prefix probabilities.
Prefix probabilities have important applications, e.g.\  
in the area of speech recognition.

Our algorithm is a minor extension
of an application of dynamic programming developed
for non-probabilistic PDTs by~\cite{LA74,BI89}, and
the treatment of probabilities is derived from~\cite{ST95}.

Assume a fixed PPDT $(\myaut,p_{\myaut})$ and a
fixed input string $a_1 \cdots a_n$. Consider a
computation of the form $c_1 \tau c_2$, where
$(\Xinit, a_1 \cdots a_i, \epsilon)$ $\pdamovesname{c_1}$
$(\alpha X, \epsilon, v_1)$,
$\tau$ is of the form 
$\myep{{\it X}}{{\it X Y'}}$, and
$(Y', a_{i+1} \cdots a_j, \epsilon)$
$\pdamovesname{c_2}$
$(Y, \epsilon, v_2)$, for
some stack symbols $X,Y',Y$,
some input positions $i$ and $j$ ($0 \leq i \leq j \leq n$),
and some output strings $v_1$ and $v_2$.
In words, the computation 
obtains top-of-stack $X$ after
scanning of $a_i$ but before scanning of $a_{i+1}$,
then applies a push transition, and then possibly
further push, scan and pop transitions, which
leads to $Y$ on top of $X$ after
scanning of $a_j$ but before scanning of $a_{j+1}$.

We now abstract away from some details of such a computation 
by just recording $X$, $Y$, $i$, $j$ and its probability 
$p_1=p_{\myaut}(c_1\tau c_2)$.
The probability $p_1$ is related to what is commonly called 
a {\em forward\/} probability, 
as it expresses the probability of the computation
from the beginning onward.%
\footnote{Forward probability as defined by \cite{ST95}
refers to the sum of the probabilities of
{\em all\/} computations from the
beginning onward that lead to a certain rule occurrence, 
whereas here we consider only one computation at a time.
We will turn to forward probabilities later in this section.}
The existence of the above computation is represented by an
object that we will call a {\em table item\/},
written as $p_1:\forward(X,Y,i,j)$.

Similarly, consider a subcomputation of the form
$\tau c_2$, where as before
$\tau$ is of the form 
$\myep{{\it X}}{{\it X Y'}}$, and
$(Y', a_{i+1} \cdots a_j, \epsilon)$
$\pdamovesname{c_2}$
$(Y, \epsilon, v_2)$, for
some stack symbols $X,Y',Y$,
some input positions $i$ and $j$ ($0 \leq i \leq j \leq n$),
and some output string $v_2$.
We express the existence of such a subcomputation
by a different kind of table item, written as
$p_2:\inner(X,Y,i,j)$, where
$p_2=p_{\myaut}(\tau c_2)$. Here, $p_2$ is related to what is commonly
called an {\em inner\/} probability, as
it expresses only the probability internally in a
subcomputation.%
\footnote{We will turn to actual inner probabilities 
later in this section.}

For technical reasons, we also need to consider
computations $c$ where
$(\Xinit, a_1 \cdots a_j, \epsilon)$ $\pdamovesname{c}$
$(Y, \epsilon, v)$, for some $Y$, $j$ and $v$.
These are represented by table items
$p_1:\forward(\bot,Y,0,j)$,
where $p_1=p_{\myaut}(c)$.
The symbol $\bot$ can be seen as an imaginary stack symbol that 
is located
below the actual bottom-of-stack element.

All table items of the above forms, and only those table items,
can be derived by the deduction system in 
Figure~\ref{f:tabular}. Deduction systems for defining
parsing algorithms have been described before by \cite{SH95};
see also \cite{SI97,SI97a} for a very similar framework.
A dynamic programming algorithm for such a deduction system
incrementally fills a {\em parse table\/} with 
table items, given a grammar and input.
During execution of the algorithm,
items that are already
in the table are matched against antecents of inference
rules. If a combination of items match all
antecents of an inference rule, then the item
that matches the consequent of that inference rule is
added to the table. This process ends when no more
new items can be added to the table.

The item in the consequent of inference rule~(\ref{e:init})
represents the fact that 
at the beginning of any computation, 
$\Xinit$ lies on top of imaginary stack
element $\bot$, no input has as yet been read, and
the product of probabilities of all transitions used 
in the represented computation is 1, since no transitions 
have been used yet.

Inference rule~(\ref{e:pushfor}) derives a table item from
an existing table item, if the second stack symbol of that 
existing item indicates that a push transition can be applied.
Naturally, the probability in the new item is the product
of the probability in the old item and
the probability of the applied transition.
Inference rule~(\ref{e:scanfor}) is very similar.

Two subcomputations are combined through a 
pop transition by inference rule~(\ref{e:popfor}),
the intuition of which can be explained as follows. 
If $W$ occurs as top-of-stack at position $i$ and 
reading the input up to $j$ results in 
$Y$ on top of $W$, and if subsequently reading the input from
$j$ to $k$ results in $X$ on top of $Y$ and
${\it YX}$ may be replaced by $Z$ by a pop transition, then
reading the input from $i$ to $k$ results in
$Z$ on top of $W$.
The probability of the newly derived subcomputation is the
product of three probabilities. 
The first is the probability of that subcomputation
up to the point where $Y$ is top-of-stack,
which is given by $p_1$; the second is the 
probability from this point onward, up to the point where
$X$ is top-of-stack,
which is given by $p_2$;
the third is the probability of the pop transition.
The second of these
probabilities, $p_2$, is defined by the inference rules for
`inner' items to be discussed next.

Inference rule~(\ref{e:pushin}) starts the investigation of
a new subcomputation that begins with a push transition.
This rule does not have any antecedents, but we may 
add an item $p_1:\forward(Z,X,i,j)$ as antecedent,
since the resulting `inner' items can only be useful for
the computation of `forward' items if at least
one item of the form $p_1:\forward(Z,X,i,j)$
exists. We will not do so
however, since this would complicate the theoretical analysis.

The next two rules, (\ref{e:scanin}) and~(\ref{e:popin}), 
are almost identical to (\ref{e:scanfor}) and~(\ref{e:popfor}).

\begin{figure}[t]
Initialization: \\[-4ex]
\tabruletwo{e:init}{
}{
1:\forward(\bot,\Xinit,0,0)
}

Push (forward): \\[-4ex]
\tabrule{e:pushfor}{
p_1:\forward(Z,X,i,j)
}{
p_1 \cdot p_{\myaut}(\tau):\forward(X,Y,j,j)
}{
\tau = \myep{X}{\it XY}
}

Scan (forward): \\[-4ex]
\tabrule{e:scanfor}{
p_1:\forward(Z,X,i,j)
}{
p_1 \cdot p_{\myaut}(\tau):\forward(Z,Y,i,j')
}{
\tau = \myscan{X}{x}{y}{\it Y} \\
(x = \epsilon \wedge j' = j)\ \vee  \\
\ \ \ (x = a_{j+1} \wedge j' = j + 1)
}

Pop (forward): \\[-4ex]
\tabrule{e:popfor}{
p_1:\forward(W,Y,i,j) \\
p_2:\inner(Y,X,j,k)
}{
p_1 \cdot p_2\cdot p_{\myaut}(\tau):\forward(W,Z,i,k)
}{
\tau = \myep{{\it Y X}}{\it Z} 
}
 
Push (inner): \\[-4ex]
\tabrule{e:pushin}{
}{
p_{\myaut}(\tau):\inner(X,Y,j,j)
}{
\tau = \myep{X}{\it XY}
}

Scan (inner): \\[-4ex]
\tabrule{e:scanin}{
p_2:\inner(Z,X,i,j)
}{
p_2 \cdot p_{\myaut}(\tau):\inner(Z,Y,i,j')
}{
\tau = \myscan{X}{x}{y}{\it Y} \\
(x = \epsilon \wedge j' = j)\ \vee  \\
\ \ \ (x = a_{j+1} \wedge j' = j + 1)
}

Pop (inner): \\[-4ex]
\tabrule{e:popin}{
p_2:\inner(W,Y,i,j) \\
p'_2:\inner(Y,X,j,k)
}{
p_2 \cdot p'_2\cdot p_{\myaut}(\tau):\inner(W,Z,i,k)
}{
\tau = \myep{{\it Y X}}{\it Z}
}

\caption{Deduction system of table items.}
\label{f:tabular}
\end{figure}

It is not difficult to see that for each complete
computation of the form
$(\Xinit, a_1 \cdots a_n, \epsilon)$ $\pdamovesname{c}$
$(\Xfinal, \epsilon, v)$, for some output string $v$,
there is precisely one derivation by the deduction system
of some table item 
$p_1:\forward(\bot,\Xfinal,0,n)$, where $p_1=p_{\myaut}(c)$.
Conversely, for each derivation of such a table item, there
is a unique corresponding computation.
Computations and derivations can be easily related to each other
by looking at the transitions in the side conditions of the
inference rules.

If follows that if we take the sum of
$p_1$ over all derivations of items
$p_1:\forward(\bot,\Xfinal,0,n)$, then we obtain
the probability assigned by $\myaut$ to the input
$w=a_1 \cdots a_n$. 

Now assume that $\myaut$ is proper and consistent. 
For a given string
$w' \in \myterm_1^\ast$, where $\myterm_1$ is the input
alphabet, we define the {\em prefix probability\/} of $w'$
to be 
$$ \sum_{w'' \in \myterm_1^\ast}\ p_{\myaut}(w' w'') $$
In other words, we sum the probabilities of all strings
$w=w'w''$ that start with prefix $w'$.
We will now show that this probability can also be expressed
in terms of the probabilities of `forward' items.

Assume that $w'= a_1 \cdots a_n$, for some $n \geq 0$.
Any computation on a string $w=w'w''$ 
that is the prefix of a complete computation 
must be of one of two types.
The first is
$(\Xinit, a_1 \cdots a_n, \epsilon)$ $\pdamovesname{c}$
$(\Xfinal, \epsilon, v)$, for some $v$, which means that
$w''=\epsilon$, so that no input beyond position $n$ needs to be
read.
The second is
$(\Xinit, a_1 \cdots a_n a_{n+1} \cdots a_m, \epsilon)$ $\pdamovesname{c_1}$
$(\alpha X, a_{n+1} \cdots a_m, v_1)$ $\pdamove{\tau}$
$(\alpha Y, a_{n+2} \cdots a_m, v_1y)$ $\pdamovesname{c_2}$
$(\Xfinal, \epsilon, v_1yv_2)$,
where $\tau$ is a scan transition
$\myscan{X}{a}{y}{Y}$ such that $a=a_{n+1}$.

The sum of probabilities of computations of the first type equals the
sum of $p_1$ over all derivations of items
$p_1:\forward(\bot,\Xfinal,0,n)$, as we have explained above.
For the second type of computation, properness and
consistency
implies that for given $c_1$ and $\tau$ as above,
the sum of probabilities of different $c_2$ must be 1.
(If that sum, say $q$, is less than $1$, then 
the sum of the probabilities of all computations cannot be
more than $1 - (1-q) \cdot p_{\myaut}(c_2) < 1$, which
is in conflict with the assumed consistency.)
Furthermore, properness implies 
that the sum of probabilities of different $\tau$ 
that we can apply for top-of-stack $X$ must be 1.
Therefore, we may conclude that
the sum of probabilities of computations of the 
second type equals the sum of $p_{\myaut}(c_1)$ over all computations
$(\Xinit, a_1 \cdots a_n, \epsilon)$ $\pdamovesname{c_1}$
$(\alpha X, \epsilon, v_1)$ such that there is at least
one scan transition of the form $\myscan{X}{a}{y}{Y}$.
This equals the sum of $p_1$ over all derivations of items
$p_1:\forward(Z,X,0,n)$, for some $Z$, such that there is at least
one scan transition of the form $\myscan{X}{a}{y}{Y}$.

Hereby we have shown how both the probability and the
prefix probability of a string can be expressed in terms of 
derivations of table items. However, the number of
derivations of table items can be infinite. The obvious
remedy lies in an alternative interpretation of the inference
rules in Figure~\ref{f:tabular}, following \cite{GO99}:
we regard objects of the form
$\forward(X,Y,i,j)$ or $\inner(X,Y,i,j)$ as
table items in their own right, and store each at most once
in the parse table.
The associated probabilities 
are then no longer those for individual derivations, 
but are the sums of probabilities
over all derivations of table items
$\forward(X,Y,i,j)$ or $\inner(X,Y,i,j)$.
Such a sum of probabilities over all
derivations of a table item is commonly
called a {\em forward\/} or {\em inner\/} probability,
respectively.

We will make this more concrete, under the assumption that
there are no cyclic dependencies, i.e.,
there is no item $\forward(X,Y,i,j)$ or $\inner(X,Y,i,j)$ that 
may occur as ancestor of itself in some derivation.
Let $T$ be the set of all items
$\forward(X,Y,i,j)$ or $\inner(X,Y,i,j)$ that can be derived
using the deduction system in Figure~\ref{f:tabular},
ignoring the probabilities.
We then define a function $p_{\tabel}$ from table items to
probabilities, as shown in Figure~\ref{f:recursive}.
We assume the function $\delta$ evaluates to 1 if its
argument is true, and to 0 otherwise.

\begin{figure}[t]
\begin{eqnarray}
\label{e:forward}
\lefteqn{p_{\tabel}(\forward(X,Y,i,j))\ =} \\
&& \delta(X = \bot \wedge Y = \Xinit \wedge 
                i = j = 0)\ + \nonumber\\
&& \delta(i=j) \cdot
\sum_{
Z, k,\tau: \atop
{ \forward(Z,X,k,i)\in T, \atop 
\tau = \myep{X}{\it XY}}
}
\hspace{-3ex} 
	p_{\tabel}(\forward(Z,X,k,i)) \cdot p_{\myaut}(\tau)\ + \nonumber\\
&& \sum_{
Z,j',x,y,\tau: \atop
{ \forward(X,Z,i,j')\in T, \atop
{ (x = \epsilon \wedge j' = j) \vee 
	(x = a_{\tiny j} \wedge j' = j - 1) , \atop
\tau = \myscan{Z}{x}{y}{\it Y} }}
}
\hspace{-8ex} 
        p_{\tabel}(\forward(X,Z,i,j')) \cdot p_{\myaut}(\tau)\ + \nonumber\\
&& \sum_{
W,Z,k,\tau: \atop
{ \forward(X,W,i,k)\in T, \inner(W,Z,k,j)\in T, \atop
\tau = \myep{\it WZ}{\it Y} }
}
\hspace{-11ex}
	p_{\tabel}(\forward(X,W,i,k)) \cdot 
	p_{\tabel}(\inner(W,Z,k,j)) \cdot p_{\myaut}(\tau) \nonumber
\end{eqnarray}
\begin{eqnarray}
\label{e:inner}
\lefteqn{p_{\tabel}(\inner(X,Y,i,j))\ =} \\
&& \delta(i=j) \cdot
\sum_{
\tau: \atop
\tau = \myep{X}{\it XY}
}
	p_{\myaut}(\tau)\ + \nonumber\\
&& \sum_{
Z,j',x,y,\tau: \atop
{ \inner(X,Z,i,j')\in T, \atop
{ (x = \epsilon \wedge j' = j) \vee
        (x = a_{\tiny j} \wedge j' = j - 1) , \atop
\tau = \myscan{Z}{x}{y}{\it Y} }}
}
\hspace{-8ex}
        p_{\tabel}(\inner(X,Z,i,j')) \cdot p_{\myaut}(\tau)\ + \nonumber\\
&& \sum_{
W,Z,k,\tau: \atop
{ \inner(X,W,i,k)\in T, \inner(W,Z,k,j)\in T, \atop
\tau = \myep{\it WZ}{\it Y} }
}
\hspace{-11ex}
        p_{\tabel}(\inner(X,W,i,k)) \cdot 
        p_{\tabel}(\inner(W,Z,k,j)) \cdot p_{\myaut}(\tau) \nonumber
\end{eqnarray}
\caption{Recursive functions to determine probabilities of
table items.}
\label{f:recursive}
\end{figure}

Each line in the right-hand sides of the two equations 
in Figure~\ref{f:recursive} can 
be seen as the backward application
of an inference rule from Figure~\ref{f:tabular}.
In other words,
for a given item, 
we investigate 
all possible ways of deriving that item as the 
consequent of different inference rules with different antecedents.
For example, the second line in the right-hand side of 
equation~(\ref{e:forward}), 
can be seen as the backward application of inference rule (\ref{e:pushfor}).

That Figure~\ref{f:recursive} is indeed equivalent to
Figure~\ref{f:tabular} follows from the fact that
multiplication distributes over addition.
If there are cyclic dependencies, then the set of equations
in Figure~\ref{f:recursive} may no longer have a closed-form
solution, but we may obtain probabilities by
an iterative algorithm that approximates the lowest non-negative 
solution to the equations \cite{ST95}.

Given the set of equations in Figure~\ref{f:recursive}
we can now express the probability of a string of length $n$ as
$p_{\tabel}(\forward(\bot,\Xfinal,0,n))$.
The prefix probability of a string of length $n$ is given by:
\begin{eqnarray}
&& p_{\tabel}(\forward(\bot,\Xfinal,0,n))\ + \\
&& \sum_{ 
X,Y,i: \atop
{ \forward(X,Y,i,n)\in T, \atop
\exists \tau,a,y,Z[\tau = \myscan{Y}{a}{y}{\it Z}] }
}
\hspace{-3ex}
p_{\tabel}(\forward(X,Y,i,n))
\end{eqnarray}

To obtain a suitable PPDT from a given PCFG,
we may apply
the strategy $\stratepLC$
from Section~\ref{s:strong}. Provided the (P)CFG is acyclic,
this strategy ensures that there are no computations of
infinite length for any given
input, which implies there are no cyclic dependencies
in the simulation of the automaton by the dynamic programming
algorithm.

Hereby we have presented a way to compute probabilities
and prefix probabilities of strings. Our approach is an alternative
to the one from \cite{JE91,ST95}, and has the advantage that
the approach is parameterized by the parsing strategy:
instead of $\stratepLC$
we may apply any other parsing strategy with the same properties
with regard to acyclic grammars.
If our grammars are even more constrained,
e.g.\ if they do not have epsilon rules, 
we may apply even simpler parsing strategies.
Different parsing strategies may differ in the efficiency
of the computation.

\section{Conclusions}

We have formalized the notion of parsing strategy as a mapping from
context-free grammars to push-down transducers, and have investigated 
the extension to probabilities. 
We have shown that the question of which
strategies can be extended to become probabilistic heavily
relies on two properties, the correct-prefix property and
the strong predictiveness property. 
The CPP is a necessary condition for
extending a strategy to become a probabilistic strategy.
The CPP and SPP together form a sufficient condition.
We have shown that there is at least one strategy 
of practical interest with the CPP but
without the SPP that cannot be extended to become a probabilistic
strategy.
Lastly, we have presented an application
to prefix probabilities.

\section*{Acknowledgements}

We gratefully acknowledge correspondence with
David McAllester,
Giovanni Pighizzini,
Detlef Prescher,
Virach Sornlertlamvanich and
Eric Villemonte de la Clergerie.

\bibliographystyle{plain}
\bibliography{/home/markjan/bib/refs}


\end{document}